\documentclass[preprint,authoryear]{elsarticle}

\usepackage[utf8]{inputenc}
\usepackage[T1]{fontenc}
\usepackage{hyperref}
\usepackage{url}
\usepackage{booktabs}
\usepackage{amsfonts}
\usepackage{nicefrac}
\usepackage{xcolor}

\usepackage{subcaption}
\usepackage{enumitem}
\usepackage{amsmath}
\usepackage{amssymb}
\usepackage{amsthm}
\usepackage{graphicx}
\usepackage{makecell}
\usepackage{xargs} 
\usepackage{float}
\usepackage{bm}
\usepackage{color}
\usepackage[ruled]{algorithm2e}

\newcommand{\vv}{\ensuremath{\bm{v}}}
\def\mW{{\bm{W}}}
\def\mP{{\bm{P}}}
\def\mI{{\bm{I}}}
\newcommand{\vect}{\text{vec}}

\newtheorem{definition}{Definition}[section]

\begin{document}

\begin{frontmatter}

\title{Mode Combinability: Exploring Convex Combinations of Permutation Aligned Models}

\author[renyi,elte]{Adrián Csiszárik\corref{c1}}
\ead{csadrian@renyi.hu}
\author[renyi,elte]{Melinda F. Kiss\corref{c1}}
\ead{mfkiss@renyi.hu}

\author[renyi]{Péter Kőrösi-Szabó}
\ead{koszpe@renyi.hu}
\author[renyi]{Márton~Muntag}
\ead{muntag@renyi.hu}
\author[renyi]{Gergely~Papp}
\ead{gergopool@renyi.hu}
\author[renyi]{Dániel~Varga}
\ead{daniel@renyi.hu}

\cortext[c1]{Corresponding author}

\affiliation[renyi]{organization={Alfréd Rényi Institute of Mathematics},
            addressline={Reáltanoda utca 13-15.}, 
            city={Budapest},
            postcode={1053}, 
            country={Hungary}}
\affiliation[elte]{organization={Eötvös Loránd University},
            addressline={Pázmány Péter sétány 1/C}, 
            city={Budapest},
            postcode={1117}, 
            country={Hungary}}

\begin{abstract}
We explore element-wise convex combinations of two permutation-aligned neural network parameter vectors $\Theta_A$ and $\Theta_B$ of size $d$. We conduct extensive experiments by examining various distributions of such model combinations parametrized by elements of the hypercube $[0,1]^{d}$ and its vicinity. Our findings reveal that broad regions of the hypercube form surfaces of low loss values, indicating that the notion of linear mode connectivity extends to a more general phenomenon which we call \emph{mode combinability}. We also make several novel observations regarding linear mode connectivity and model re-basin. We demonstrate a transitivity property: two models re-based to a common third model are also linear mode connected, and a robustness property: even with significant perturbations of the neuron matchings the resulting combinations continue to form a working model. Moreover, we analyze the functional and weight similarity of model combinations and show that such combinations are non-vacuous in the sense that there are significant functional differences between the resulting models.
\end{abstract}


\end{frontmatter}

\section{Introduction}
\label{sec:intro}

Linear mode connectivity (LMC) is an extensively studied phenomenon associated with deep neural networks. The possibility of interpolating trained models in the weight space while maintaining a relatively low loss value is an intriguing property, whose understanding is of both theoretical and practical interest. It offers many valuable insights into the structure of the loss surfaces and the underlying regularities of deep learning models \citep{entezari2022permutation, ainsworth2022rebasin}. Since its discovery \citep{frankle2020linear}, it has influenced several research directions, including model merging \citep{NEURIPS2020_Singh_Jaggi}, federated learning \citep{jordan2022repair}, model pruning \citep{frankle2020linear}, or continual learning \citep{mirzadeh2021linear}. In a broader context, the motivation to explore LMC stretches to recent advancements regarding changing, fine-tuning, or merging knowledge in foundation models, as several related methods \citep{hu2022lora, dettmers2023qlora, meng22_loced} can also be interpreted as additive operations in the weight space with the aim of maintaining or aggregating model performance. Delving into LMC is a step toward understanding how model combinations operating in the weight space relate to model functionality.

The primary context of the present work is given by the results of \cite{ainsworth2022rebasin}. They present empirical evidence --- confirming a hypothesis formerly formulated by \cite{entezari2022permutation} --- that a neural network loss landscape often has a single basin which is fundamentally convex, after accounting for all possible permutation symmetries of hidden units. They introduce Git Re-Basin, a technique to identify relevant permutations of neurons, and then linearly interpolate between the `permutation-aligned' models (i.e., demonstrate LMC) to empirically show the approximate convexity of the resulting loss basin.

Our work aims to provide novel insights on this phenomenon by studying it within a framework of a more general model composition approach. More precisely, instead of linearly interpolating between model $\Theta_A$ and $\Theta_B$ by taking convex combinations $\lambda \cdot \Theta_{A} + (1-\lambda) \cdot \Theta_{B}$ parametrized by a single scalar $\lambda \in [0,1]$, we explore the possibility of forming low-loss regions by taking the convex combinations \emph{element-wise} in the weight space. We parametrize such combinations by the points of the hypercube $[0,1]^d$, where $d$ is the number of parameters, i.e., in our approach, each parameter has a separate interpolation coefficient. Such extension of parametrization introduces new degrees of freedom that enable us to view the loss landscape from a broader perspective and pave the way for novel model composition approaches.

\medskip
We summarize our main contributions as follows:
\begin{itemize}[noitemsep]
    \item We experimentally demonstrate that linear mode connectivity is a special case of a more general phenomenon we call \emph{mode combinability}. More precisely, we show that broad regions of the space of element-wise convex combinations of re-basined model pairs exhibit accuracy and loss close to the original models.
    
    \item We empirically observe novel properties of LMC in the context of re-basin. Among these properties are \emph{transitivity}, where two models $\pi_B(\Theta_{B})$ and $\pi_C(\Theta_{C})$ re-based to the basin of $\Theta_A$ are also linear mode connected; and a \emph{robustness} to perturbations of neuron pair matching.
    
    \item We investigate the functional and weight dissimilarities of model combinations to provide a more detailed picture than merely considering model accuracies. We demonstrate the non-vacuousness of model combinations by highlighting the differences between the combined models and the original ones.
\end{itemize}

The remainder of the paper is organised as follows. Section~\ref{sec:related_work} presents an overview of related work. Section~\ref{sec:preliminaries} provides the necessary preliminaries, introduces the element-wise combinations, and the tools for our investigation formally. Section~\ref{sec:combinations_in_the_hyperrectangle} exposes the experimental exploration of element-wise convex combinations. In Section~\ref{sec:combinations_beyond_the_hyperrectangle}, we go beyond convex combinations and study parametrizations passing the boundary of the hypercube. This section also covers the transitivity property of LMC, a study of combining three models, and demonstrates model combinations that outperform the originals. In Section~\ref{sec:func_and_weight_similiarity}, we investigate functional and weight differences of model combinations, and explore the robustness of model alignment to perturbations. Finally, conclusions are presented in Section~\ref{sec:conclusion}.

\section{Related work}
\label{sec:related_work}

\subsection{Linear mode connectivity and model re-basing}

The first mode connectivity observations \citep{draxler2018essentially, garipov2018loss} were looking at piece-wise linear mode connecting paths. Linear mode connectivity was first demonstrated by \citet{frankle2020linear} in the context of the Lottery Ticket Hypothesis focusing specifically on networks that were trained from the same initialization. \cite{entezari2022permutation} conjectured linearly mode connected SGD solutions even for networks that have been trained from different initializations. A crucial ingredient of their approach is that one should take into account the permutation symmetries of neural networks. The work of \cite{ainsworth2022rebasin} then realized such permutations layer-wise with a simple greedy neuron matching algorithm family dubbed Git Re-Basin. The results in \cite{ainsworth2022rebasin} were demonstrated only on networks using Layer Normalization \citep{ba2016layer} and sufficiently wide networks (16x or 32x wider than the regular baseline). It was \cite{jordan2022repair} who extended their results to Batch Normalized \citep{ioffe2015batch} networks, and networks that have been trained without normalization --- thus, significantly weakening the necessary conditions concerning normalization for the Git Re-Basin algorithm to work.

\subsection{Model merging and knowledge aggregation in foundation models} An important related area is model merging \cite{mcmahan2017communication, NEURIPS2020_Singh_Jaggi, matena2022merging}, where the objective is to amalgamate the knowledge captured in distinct models by operating on the network weights. The findings related to LMC suggest at first glance an unorthodox and surprisingly simple methodology to accomplish this task: by taking the average of model weights. The strangeness of this approach stems from the contrast between the simplicity of the addition operation performed in the weight space and the potential complexity of the task of aligning and extending the generalization capabilities of deep learning models. Recent advancements regarding changing, fine-tuning, or merging knowledge in foundation models \citep{hu2022lora, dettmers2023qlora, meng22_loced} can also be interpreted as additive operations in the weight space. These approaches add low-rank residuals with a the aim of maintaining of extending functionality.

\section{Preliminaries and notation}
\label{sec:preliminaries}

Let $A$ and $B$ denote two models which have identical architectures with $d \in \mathbb{N}$ parameters but were trained from different random initializations and batch ordering. Let $\Theta_A \in \mathbb{R}^d$ and $\Theta_B \in \mathbb{R}^d$ denote the weights of the two models after being trained to convergence. With a slight abuse of notation, we will identify models with their weight vectors and, e.g., refer as model $\Theta_A$ and model $\Theta_B$ to the model $A$ and $B$ having their parameters set to $\Theta_A$ and $\Theta_B$, respectively.

Linear mode connectivity and the approximate convexity of the loss basin is captured through the notion of loss barrier.

\begin{definition}[Loss barrier \citep{frankle2020linear}]
Given two models $\Theta_A$ and $\Theta_B$, such that $\mathcal{L}(\Theta_A) \approx \mathcal{L}(\Theta_B)$, the loss barrier is defined as $\max_{\lambda \in [0,1]} \mathcal{L}((1-\lambda) \Theta_A + \lambda \Theta_B) - \frac{1}{2}(\mathcal{L}(\Theta_A) + \mathcal{L}(\Theta_B))$, where $\mathcal{L}$ is the utilized loss function.
\end{definition}

Two networks are considered linear mode connected if the loss barrier between them is small (zero, or near zero). \cite{ainsworth2022rebasin} demonstrates zero-barrier linear mode connectivity between $\Theta_A$ and $\pi(\Theta_B)$, where $\pi(\Theta_B)$ is a weight vector obtained by permuting the neurons of $\Theta_B$. The permutation leaves the network $\pi(\Theta_B)$ functionally identical to $\Theta_B$, while `re-basing' it to the basin of $\Theta_A$. This can be perceived as establishing an alignment between the two models, thus, we use the term `aligned' for such $\Theta_A$ and $\pi(\Theta_B)$ model pairs. The permutation is obtained by an iterative greedy matching algorithm of which they present three variants (matching activations, weights, or learning with a straight-through estimator). We treat these algorithms as black-boxes. For completeness, we reiterate the weight matching algorithm in ~\ref{appendix:algorithm}, but refer to \cite[Section 3]{ainsworth2022rebasin} for further details.

In the present work, we extend the space of possible model combinations from convex combinations to \emph{element-wise} convex combinations.

\begin{definition}[Element-wise convex combination]
For a coefficient vector $\vv\in [0,1]^d$, let $\Theta_{\vv} \in \mathbb{R}^d$ be an element-wise convex combination of $\Theta_A$ and $\Theta_B$ with coefficients $\vv$ if
\begin{equation}
    \Theta_{\vv} = \vv \odot \Theta_A + (\bm{1} - \vv) \odot \Theta_B,
\end{equation}
where $\vv = (v_1, \dots, v_d)\in \mathbb{R}^d$, $\bm{1} \in \mathbb{R}^d$ is the all-ones vector, and $\odot$ is the element-wise product.
\end{definition}

So for every element $\Theta_{A_i}\in \Theta_A$ and $\Theta_{B_i}\in \Theta_B$ ($i\in [d]$), we can choose the coefficients $v_i$ of the convex combination independently. (With this notation, the experiments in \cite{ainsworth2022rebasin} belong to the special case when $\vv = \bm{\alpha} = (\alpha, \dots, \alpha)$ for an $\alpha \in [0,1]$, and $\Theta_B = \pi(\Theta_B)$, where $\pi$ is the permutation found by the model aligning procedure.) The element-wise convex combination of model parameters $\Theta_A$ and $\Theta_B$ determine a hyperrectangle which is thus parametrized by the hypercube.

We often deal with samples from model distributions supported on this hyperrectangle (or beyond), so we extend the definition of loss barrier accordingly. For a sample from a model combination distribution, we define the \emph{empirical loss barrier} by comparing the performance of the worst-performing model in the sample with the average performance of the original models the samples were deduced from.

\begin{definition}[Empirical loss barrier on a sample] 
Given two models $\Theta_A$ and $\Theta_B$, such that $\mathcal{L}(\Theta_A) \approx \mathcal{L}(\Theta_B)$, and a sample $\{\vv_1, \vv_2, \dots, \vv_n\}$ of size $n \in \mathbb{N}$ from a distribution $p(\vv)$ of model combination parametrizations supported on $\mathbb{R}^d$, the empirical loss barrier of this sample is defined as $\max_{i \in [n]} \mathcal{L}(\Theta_{\vv_i}) - \frac{1}{2}(\mathcal{L}(\Theta_A) + \mathcal{L}(\Theta_B))$, where $\mathcal{L}$ is the utilized loss function, and $\Theta_{\vv_i}$ is the combined model formed with the element-wise combination coefficient vector $\vv_i$.
\end{definition}

We define the \emph{empirical accuracy barrier} on a sample of model combinations analogously, by subtracting the worst-performing model in terms of accuracy from the average accuracy of the two original models.

\section{Model combinations in the hyperrectangle}
\label{sec:combinations_in_the_hyperrectangle}

A complete empirical exploration of element-wise convex combinations is infeasible even for small networks. Thus, we conduct our investigation instead by taking notable and meanwhile feasible distributions of model combinations, and investigate how they behave under the selected conditions.

\subsection{The general experiment setup}

We conduct our experiments with two different vision architectures: ResNet-20 \citep{he2016deep} and a simple non-residual convolutional network called Tiny-10 \citep{kornblith2019similarity}. We train both networks on the CIFAR-10 dataset \citep{krizhevsky2009learning}, starting from different initializations and using different batch orders. We follow rather standard training methodologies which (along with the precise architectural descriptions) are detailed in \ref{appendix:training_details}. For a more compact exposition, in the main text of this paper the figures are mostly correspond to ResNet-20, and the results for the Tiny-10 network can be found in \ref{appendix:additional_figures}.
 
We follow the filter weight matching algorithm presented in \cite{ainsworth2022rebasin} to obtain an aligning permutation $\pi$. (The three algorithms presented in their paper are very close in performance.) The baseline model width is set in such a way that the first convolutional layer has 16 filters. For our experiments, we use models with a width multiplier of 32 if not noted otherwise --- a width where model re-basing already works in a stable manner.

For the sake of comparison we often show the results for combinations between original (`naïve') models, not just between aligned (`permuted') models.

\subsection{Sampling from the unit hypercube}
\label{sec:sampling}

\begin{figure}
  \centering
  \begin{subfigure}[b]{0.19\linewidth}
    \centering
    \includegraphics[width=\linewidth]{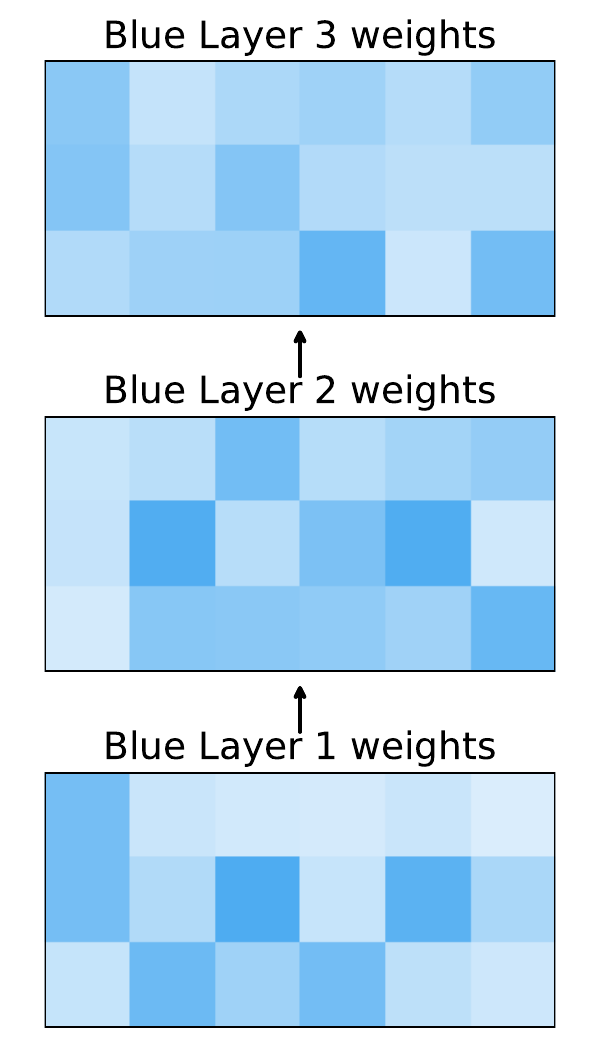}
    \caption{Blue model}
  \end{subfigure}
  \begin{subfigure}[b]{0.19\linewidth}
    \centering
    \includegraphics[width=\linewidth]{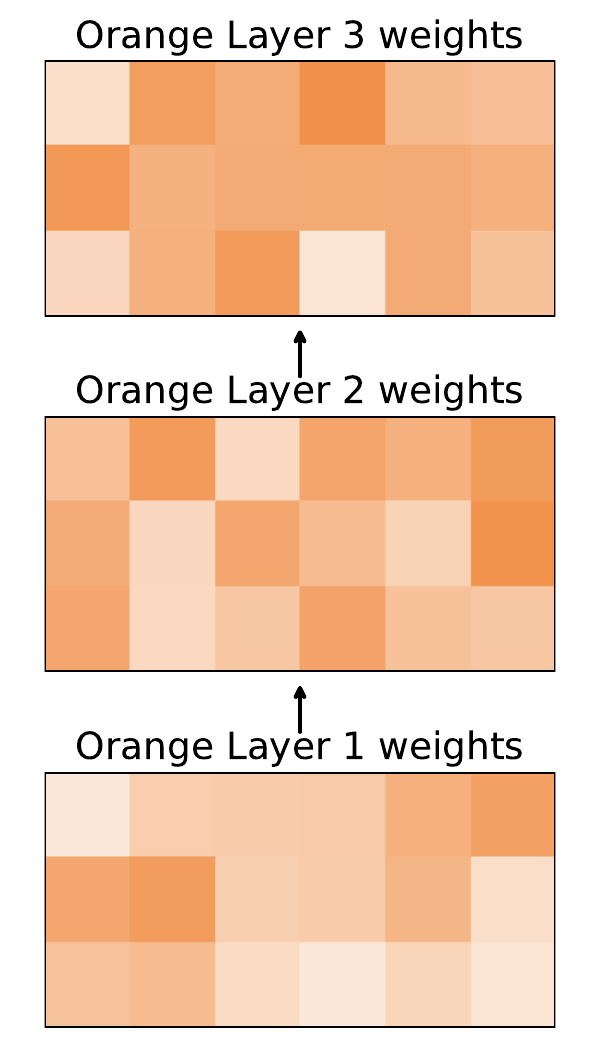}
    \caption{Orange model}
  \end{subfigure}
  \begin{subfigure}[b]{0.19\linewidth}
    \centering
    \includegraphics[width=\linewidth]{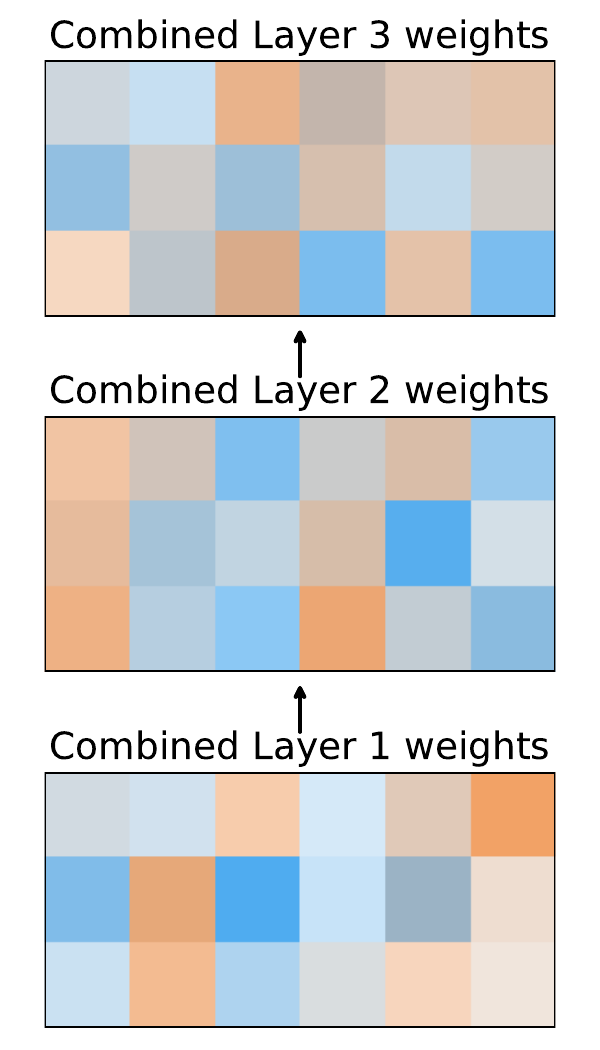}
    \caption{Uniform}
  \end{subfigure}
  \begin{subfigure}[b]{0.19\linewidth}
    \centering
    \includegraphics[width=\linewidth]{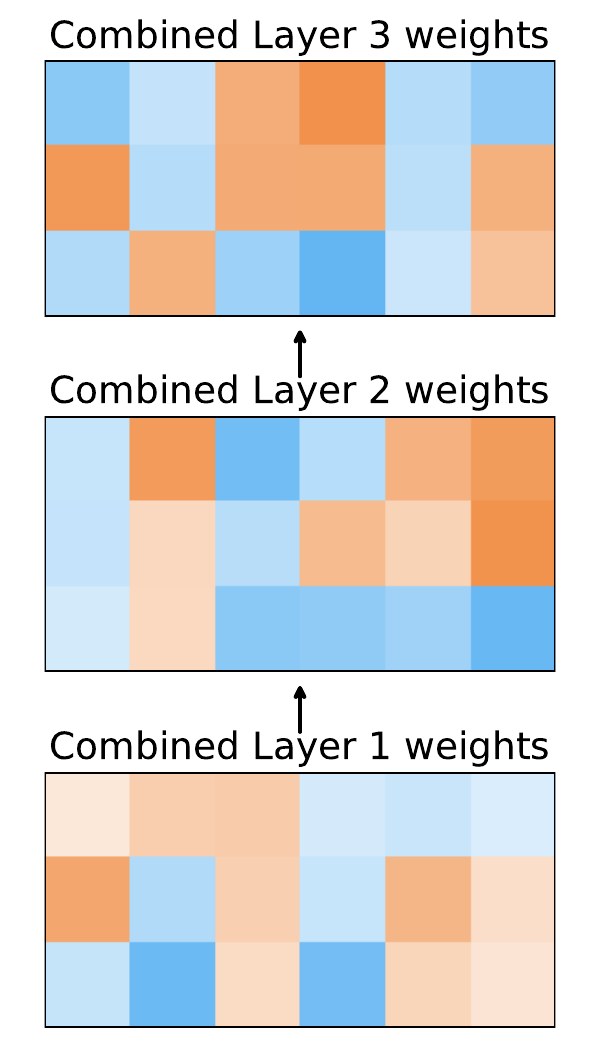}
    \caption{Bernoulli}
  \end{subfigure}
  \begin{subfigure}[b]{0.19\textwidth}
    \centering
    \includegraphics[width=\linewidth]{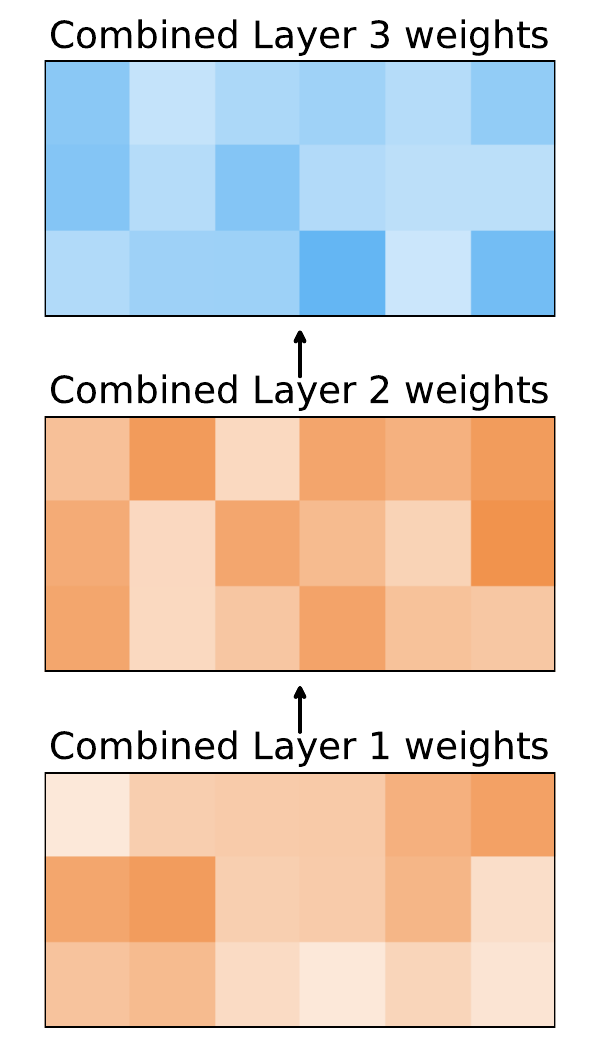}
    \caption{Stitching}
  \end{subfigure}

  \caption{Schematic figures of combinations of a `blue' and an `orange' model. Colored rectangles depict the layers of the combined networks, in which each square cell represents a weight obtained by mixing the corresponding weights of the two models. The weight value is represented by the brightness of the cell, and the mixing is illustrated by interpolating between the colors of the blue and the orange models. For this illustration, the weights are distributed uniformly.}
  \label{fig:schematic_combinations}
\end{figure}

In this section, we look at different types of distributions on the unit cube $[0,1]^d$. We use samples from these distributions as parametrizations of element-wise convex model combinations.
Most of the distributions we used are designed in such a way, to have a real-valued parameter that serves a similar role to an `interpolation coefficient'. Figure~\ref{fig:schematic_combinations} illustrates schematically some of the model combinations presented in this section.

\subsubsection{Sampling from the uniform distribution on the unit hypercube}

\begin{figure}
\centering
    \subfloat[Loss]{%
        \includegraphics[trim=0cm 0cm 0cm 0cm, clip=true, width=0.5\textwidth]{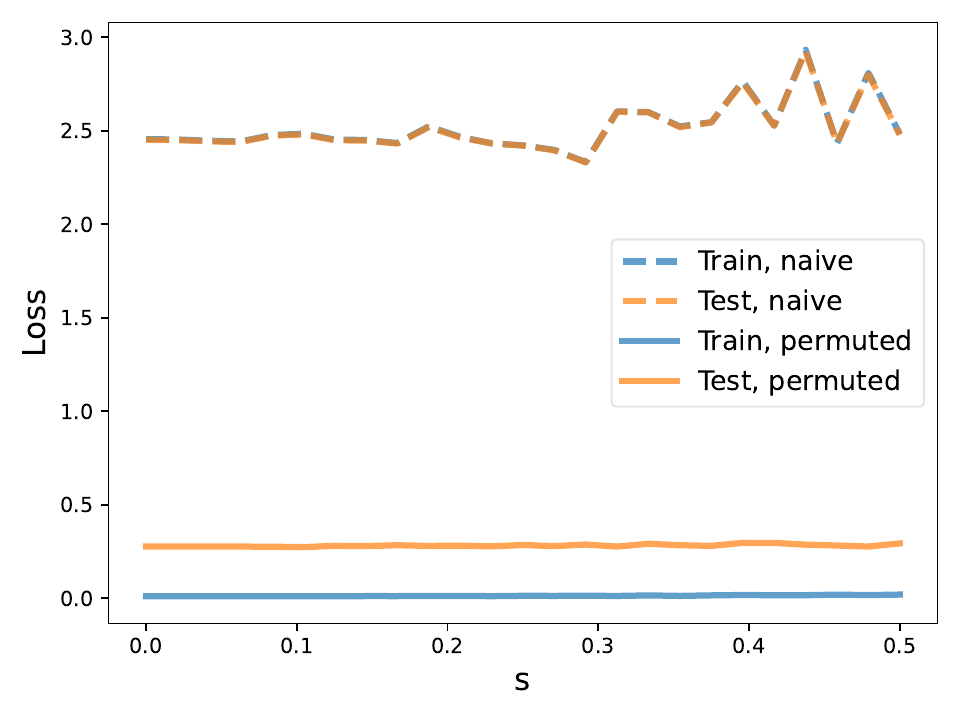}%
        \label{fig:uniflossresnet}%
        }%
    \hfill%
    \subfloat[Accuracy]{%
        \includegraphics[trim=0cm 0cm 0cm 0cm, clip=true, width=0.5\textwidth]{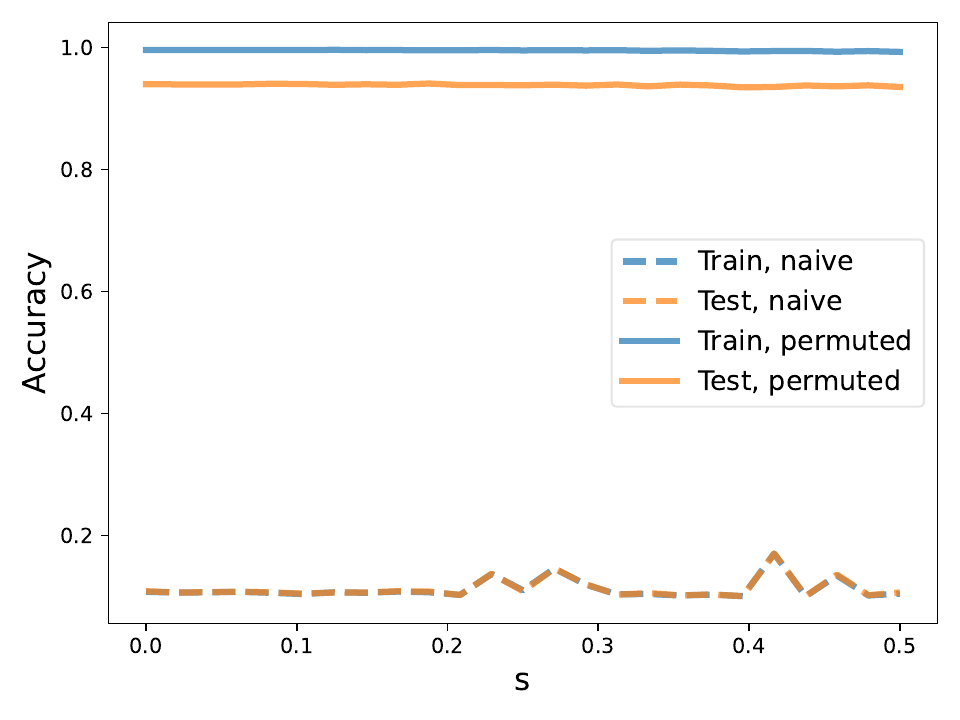}%
        \label{fig:unifaccresnet}%
        }%
    \caption{Performance of ResNet-20 model combinations corresponding to uniform sampling of element-wise coefficients between $[0.5-s, 0.5+s]$. Horizontal axes denote $s$ (with $25$ values chosen equidistantly), vertical axes denote performance in terms of (a) loss and (b) accuracy.}%
    \label{fig:uniformresnet}%
\end{figure}

The first distribution we experiment with is the uniform distribution on the unit cube (corresponding to model combinations where each weight gets its own interpolation coefficient independently and uniformly). In Figure~\ref{fig:uniformresnet} we depict model results with uniform distribution between $[0.5-s, 0.5+s]$ for various $s \in [0, 0.5]$ (note that $s=0.5$ corresponds to the uniform sampling from $[0,1]^d$). We observe that all the sampled model combinations attain high accuracy and low loss values, with empirical loss barrier 0.044 and empirical accuracy barrier 0.012.

\subsubsection{Extending the interpolation: uniform distribution on a smaller cube}

\begin{figure}
        \centering
        \subfloat[Loss]{%
            \includegraphics[trim=0cm 0cm 0cm 0cm, clip=true, width=0.5\textwidth]{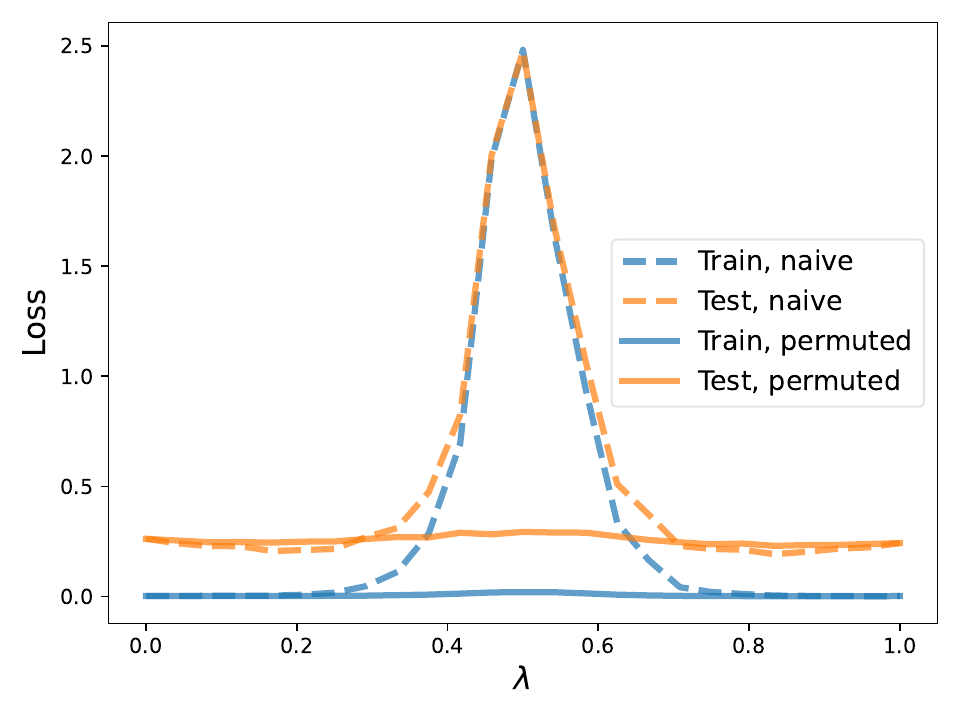}%
            \label{fig:cube_intersect_resnet_loss}%
            }%
        \hfill%
        \subfloat[Accuracy]{%
            \includegraphics[trim=0cm 0cm 0cm 0cm, clip=true, width=0.5\textwidth]{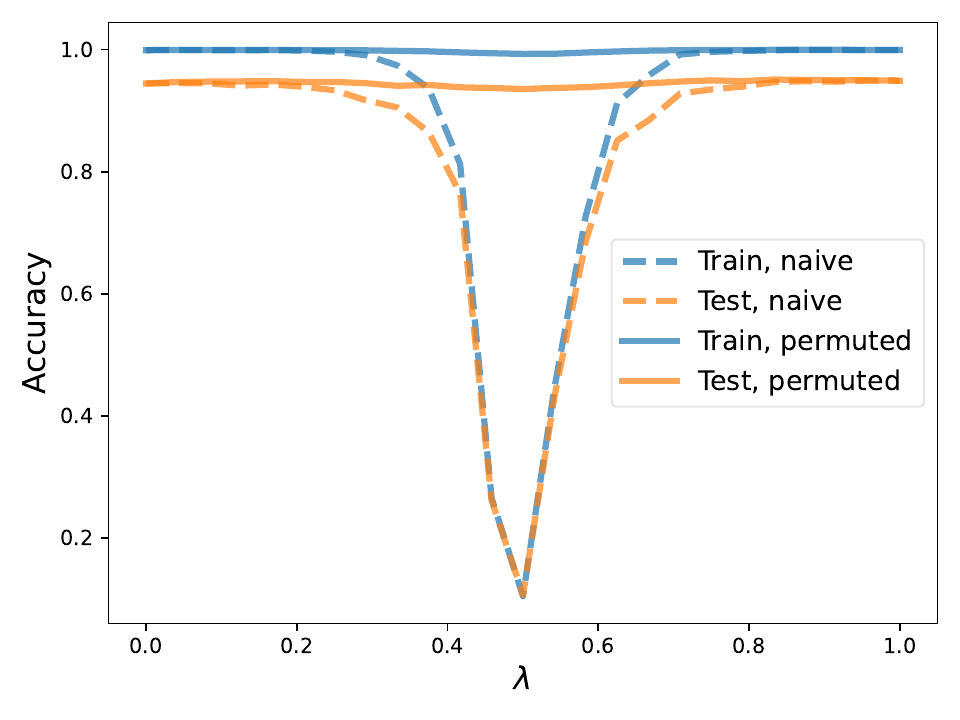}%
            \label{fig:cube_intersect_resnet_acc}%
            }%
        \caption{Performance of ResNet-20 model combinations with coefficients sampled from smaller cubes as described in Section~\ref{sec:sampling}. Horizontal axes denote the center of the small hypercube (with $25$ values chosen equidistantly), vertical axes denote performance in terms of (a) loss and (b) accuracy.}
        \label{fig:cube_intersect_resnet}
\end{figure}

We now extend the single parameter linear interpolation with an approach that still has a single parameter and retains the two original models as endpoints, however, it differs in that it samples from smaller cubes when interpolating. Here, we again have a parameter $\lambda \in [0,1]$, which controls which subcube of $[0,1]^d$ we sample from uniformly. Namely, if $\lambda \le 0.5$, we sample uniformly from the cube $[0, 2\lambda]^d$, and if $\lambda > 0.5$, we sample uniformly from the cube $[2\lambda -1, 1]^d$. (Note that for $\lambda = 0$ we get model A, and for $\lambda = 1$ we get model B.) We again choose $25$ numbers equidistantly from $[0, 1]$ for the parameter $\lambda$, and plot the loss and accuracy of the combined models in Figure~\ref{fig:cube_intersect_resnet}. We observe low barrier values and high accuracy, with empirical loss barrier 0.041 and empirical accuracy barrier 0.011.

\subsubsection{Uniform distribution on the intersection of the unit cube and a hyperplane}

Here we present another natural method of interpolating between the two models, by sweeping through the hyperrectangle with a hyperplane.

We want to sample uniformly from the intersection of the hyperplane $\{\bm{x} \in \mathbb{R}^d : \frac{1}{d} \sum_{i=1}^d x_i = \alpha \}$ and $[0,1]^d$, for some parameter $\alpha \in [0,1]$. This polytope, denoted by $P(d, \alpha)$, is hard to sample from. Hence, we instead sample from another distribution that is a good approximation of it.

Taking $d$ i.i.d. instances of any one-dimensional exponential distribution $f(x; \lambda)$, the density function of the joint distribution is constant on any hyperplane $\frac{1}{d} \sum_{i=1}^d x_i = \alpha$. Hence, over $[0, 1]^d$, the same is true for $\hat{f}(x;\lambda)$, the same distribution truncated to $[0, 1]$. This suggests the following procedure: when given the task of uniformly sampling from $P(d, \alpha)$, we instead choose a $\lambda$ such that $\mathbb{E}[\hat{f}(x;\lambda)] = \alpha$, and pick $d$ i.i.d. samples from $\hat{f}(x;\lambda)$. This procedure gives identical results to ``mishearing'' $\alpha$, and sampling from $P(d, \alpha + \varepsilon)$ instead of the required $P(d, \alpha)$, for some $\varepsilon$ random noise. The central limit theorem guarantees that the magnitude of this $\varepsilon$ noise is $O(\frac{1}{\sqrt{d}})$ with high probability. The appropriate $\hat{f}(x;\lambda)$ can be found by numerical optimization; incidentally, it is the maximum entropy distribution with support $[0, 1]$ and mean $\alpha$.

\begin{figure}
    \centering
    \subfloat[Loss]{%
        \includegraphics[trim=0cm 0cm 0cm 0cm, clip=true, width=0.5\textwidth]{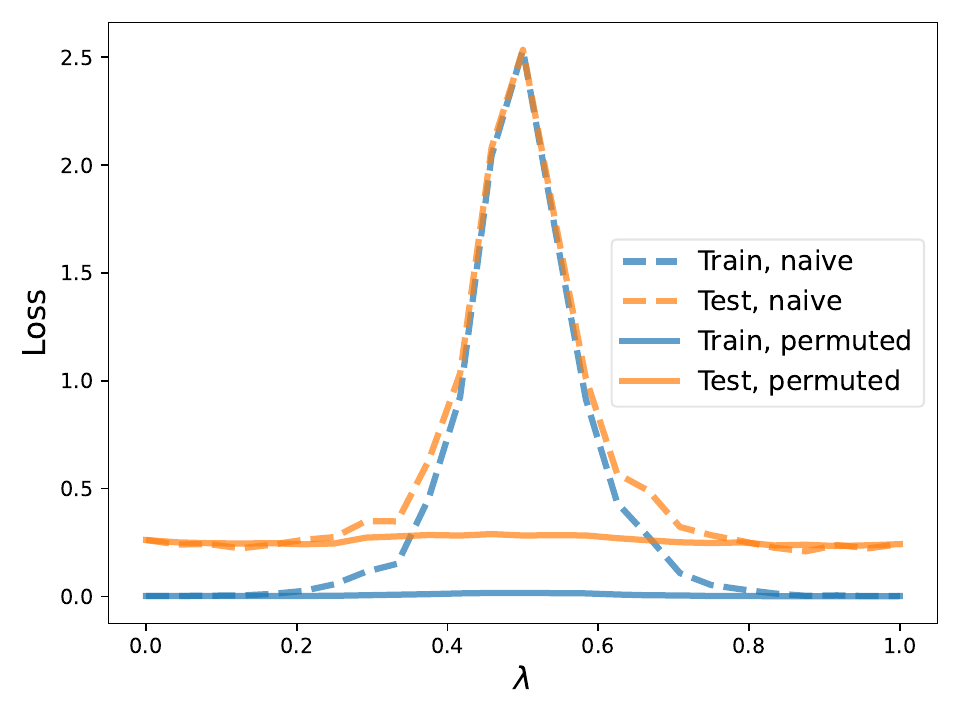}%
        \label{fig:plane_intersect_resnet_loss}%
        }%
    \hfill%
    \subfloat[Accuracy]{%
        \includegraphics[trim=0cm 0cm 0cm 0cm, clip=true, width=0.5\textwidth]{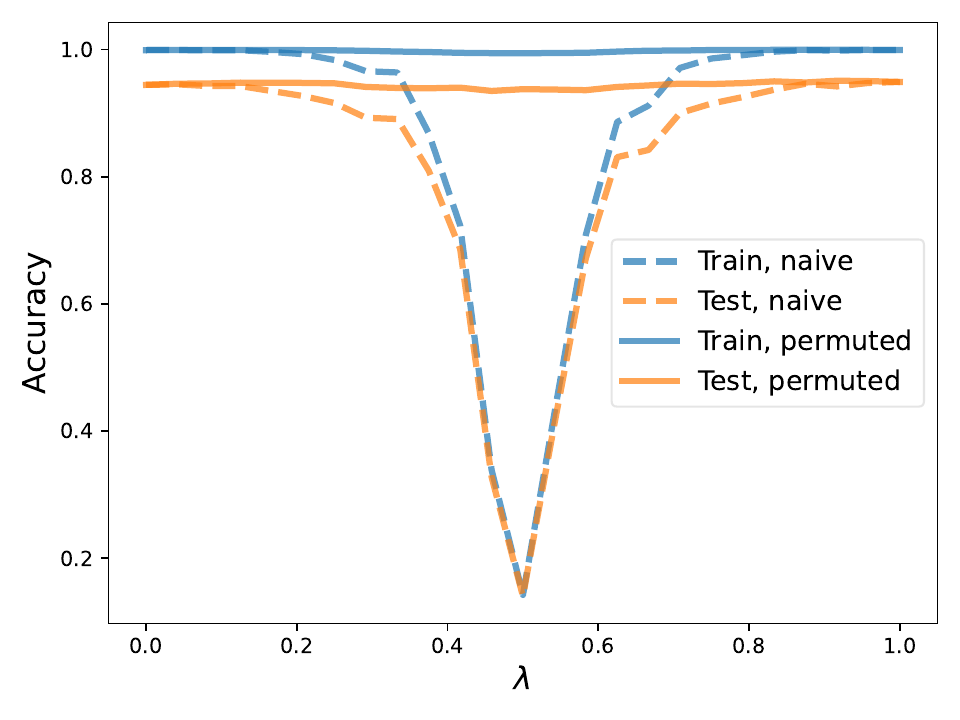}%
        \label{fig:plane_intersect_resnet_acc}%
        }%
    \caption{Performance of ResNet-20 model combinations for uniform distribution on the hyperplane described in Section~\ref{sec:sampling}.}
    \label{fig:plane_intersect_resnet}
\end{figure}

The results are shown in Figure~\ref{fig:plane_intersect_resnet}. Again, we see model combinations that expand the range of well-performing models (empirical loss barrier: 0.037, empirical accuracy barrier: 0.012).

\subsubsection{Sampling the vertices: Bernoulli distribution}

\begin{figure}
    \centering
    \subfloat[Loss]{%
        \includegraphics[trim=0cm 0cm 0cm 0cm, clip=true, width=0.5\textwidth]{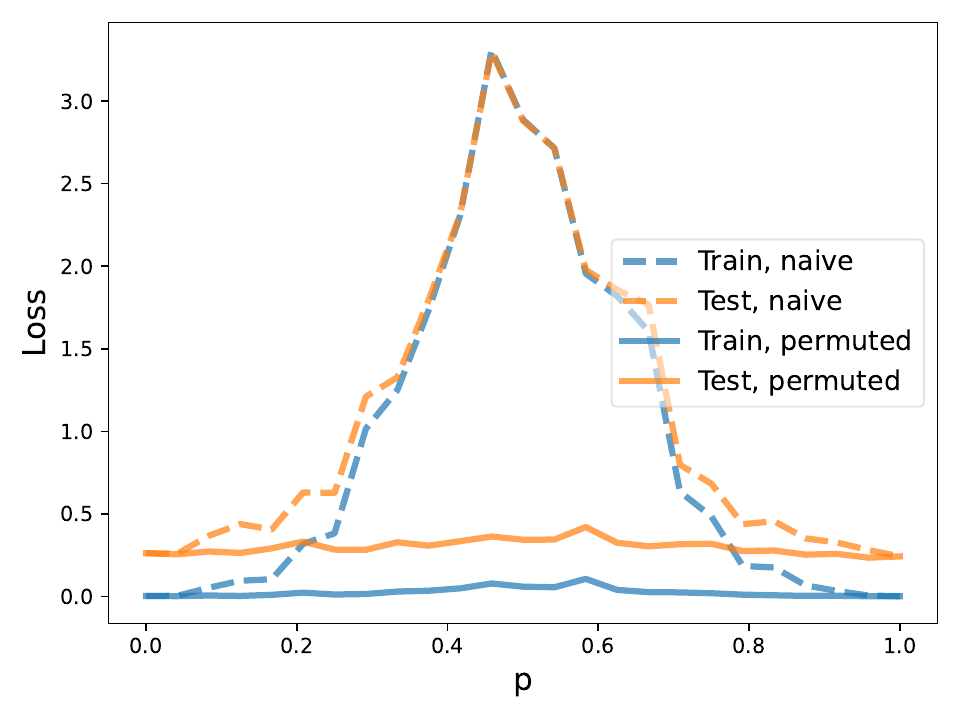}%
        \label{fig:bernoulli_resnet_loss}%
        }%
    \hfill%
    \subfloat[Accuracy]{%
        \includegraphics[trim=0cm 0cm 0cm 0cm, clip=true, width=0.5\textwidth]{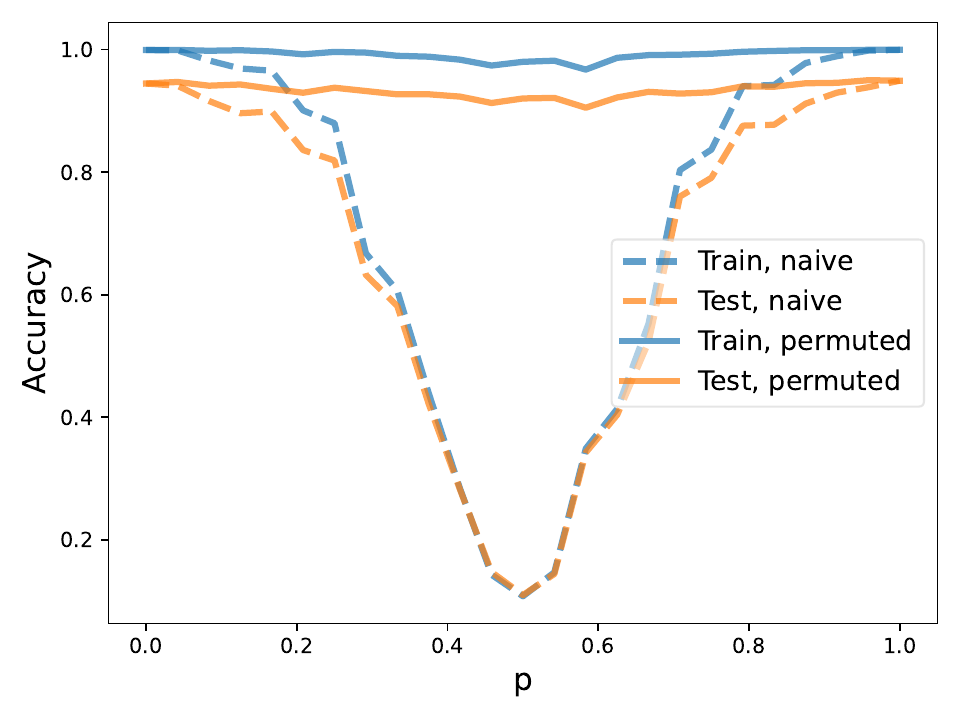}%
        \label{fig:bernoulli_resnet_acc}%
        }%
    \caption{Performance of ResNet-20 model combinations for the Bernoulli distribution on the cube. Horizontal axes denote the Bernoulli parameter $p$ (with $25$ values chosen equidistantly).}
    \label{fig:bernoulli_resnet}
\end{figure}

Next, we sample the coordinates of the $\vv$ vector independently from a Bernoulli distribution with parameter $p \in [0,1]$. This refers to model combinations where each weight of the combined network is equal to either the corresponding weight of $\Theta_A$ or $\Theta_B$. Figure~\ref{fig:bernoulli_resnet} presents the results for different parameter values of the Bernoulli distribution represented on the horizontal axis. We again observe that all model combinations result in high-performing models (empirical loss barrier: 0.168, empirical accuracy barrier: 0.042). 

\subsection{Aligned models are stitchable with identity stitching}

\begin{figure}
\centering
    \subfloat[Loss]{
        \includegraphics[trim=0cm 0cm 0cm 0cm, clip=true, width=0.49\linewidth]{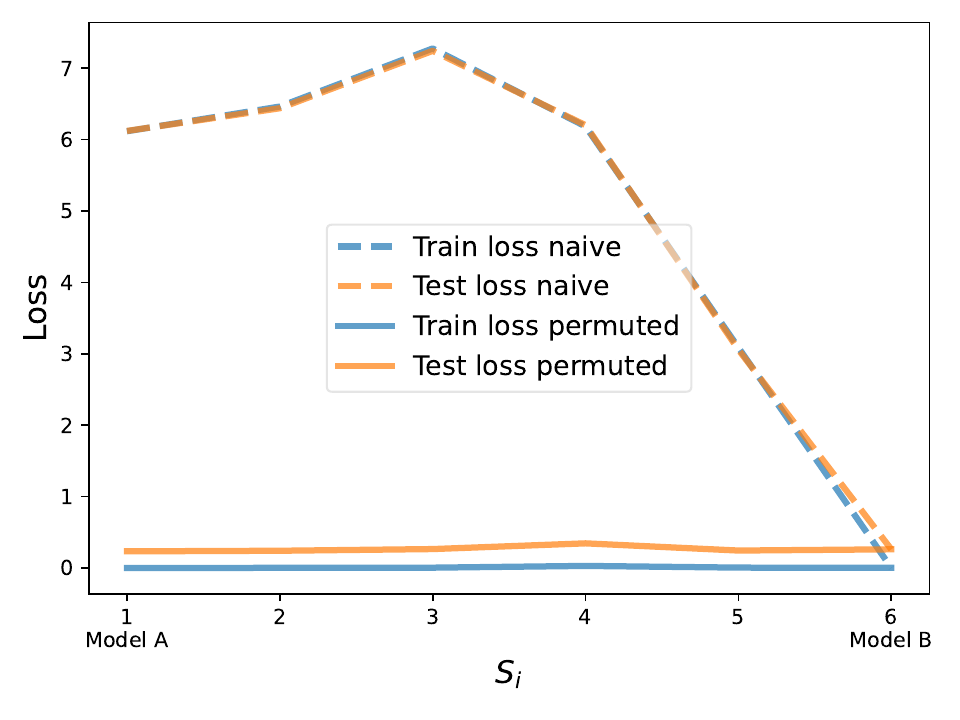}
    }%
    \subfloat[Accuracy]{
        \includegraphics[trim=0cm 0cm 0cm 0cm, clip=true, width=0.49\linewidth]{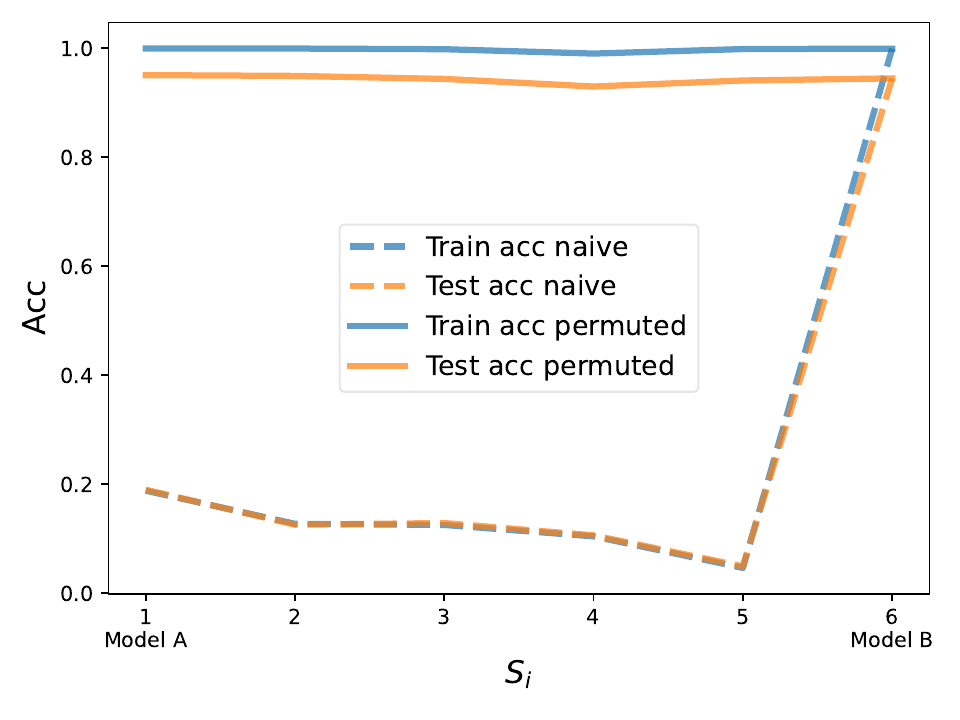}
    }
    \caption{ResNet-20 identity stitching. Horizontal axes denote the layer where the stitching is realised, vertical axes denote performance in terms of (a) loss and (b) accuracy. }
    \label{fig:stitching_resnet}
\end{figure}

Another notable set of hypercube vertices corresponds to model stitching \citep{lenc_vedaldi_19,csiszarik2021similarity,bansal2021revisiting}, where one wants to combine the lower part of model $\Theta_A$ with the upper part of model $\Theta_B$ with a stitching map making the conversion between the two representation spaces. The stitching map is usually constrained to have low complexity, e.g., be an affine transformation. In our parametrization, for a given layer $l \in \{1, \dots, L\}$ in an $L$ layered network, we set the coefficients to 0 for each layer $i \leq l$ and 1 for layers $i > l$. Such coefficients correspond to model stitching of model $\pi(\Theta_B)$ and $\Theta_A$ with the identity stitching map. Figure~\ref{fig:stitching_resnet} depicts the results. We observe that all resulting configurations attain high accuracy and low loss. Therefore, while it is necessary to have networks wide enough for the re-basin algorithm to work, with that premise, a successful stitching between $\Theta_A$ and $\Theta_B$ (note the absence of $\pi$) can be achieved using only permutation matrices --- functions of much lower complexity than arbitrary affine transformations.

\subsection{Discussion}

\begin{figure}
\centering
    \subfloat[Loss]{
        \includegraphics[trim=0cm 0cm 0cm 0cm, clip=true, width=0.49\linewidth]{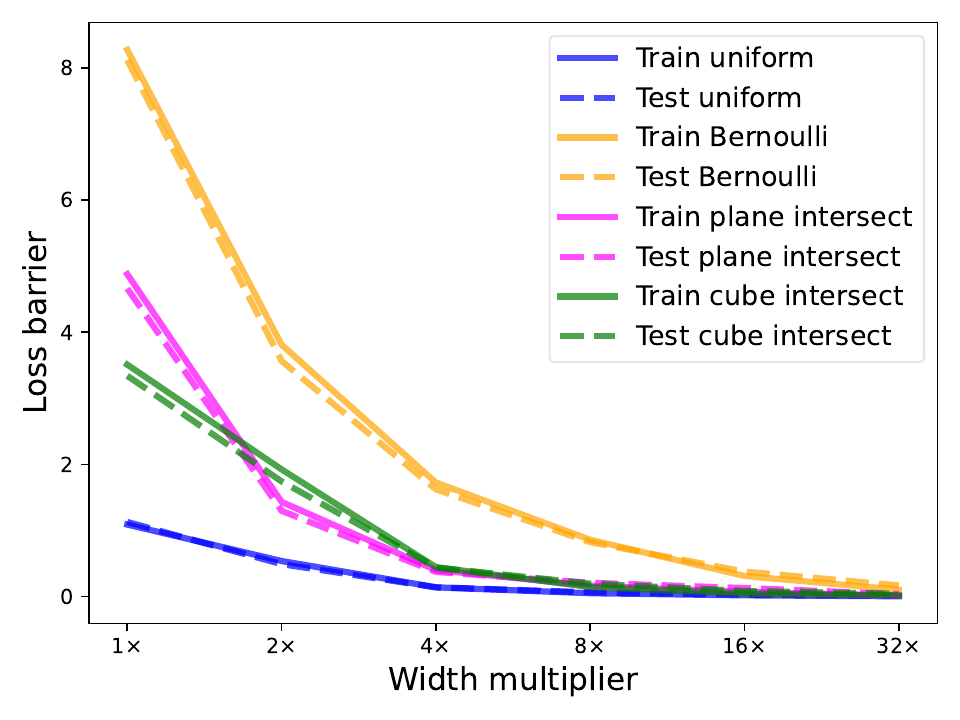}
    }%
    \subfloat[Accuracy]{
        \includegraphics[trim=0cm 0cm 0cm 0cm, clip=true, width=0.49\linewidth]{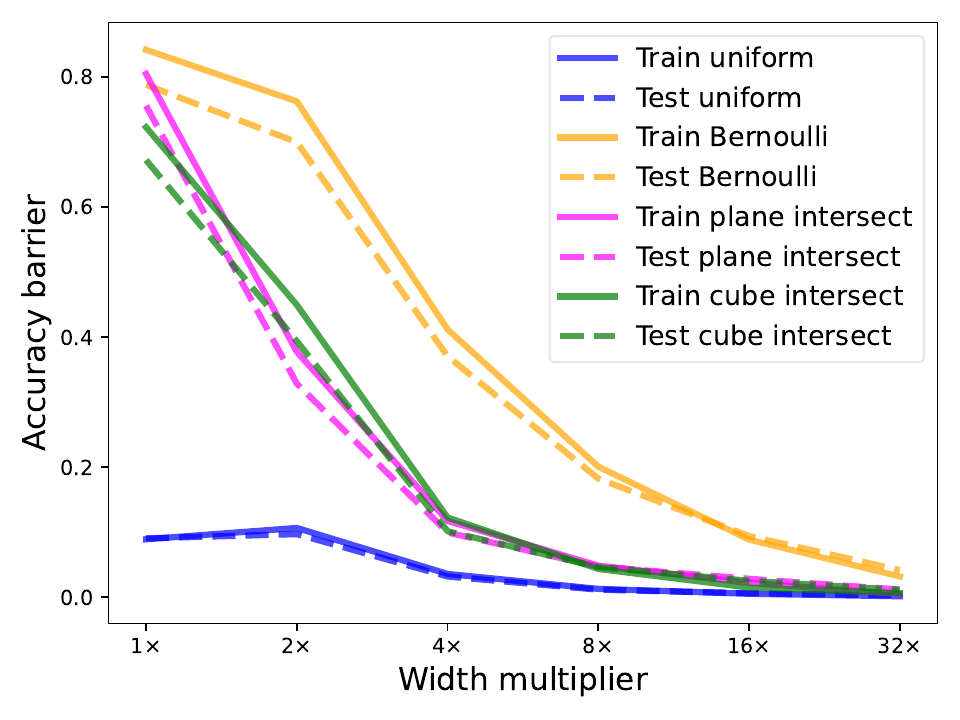}
    }%
    \caption{Different model combinations and their performance with different network widths. Network is ResNet-20. Plotting performance in terms of (a) loss barrier and (b) accuracy barrier (i.e., the worst performing model is depicted for each sampling scheme and width multiplier).}
    \label{fig:resnet20_width_plots}
\end{figure}

Figure~\ref{fig:resnet20_width_plots} presents the role of network width for several of the above sampling schemes. (Note that while in this figure each presented combination method involves a real parameter we use to interpolate, the underlying sampling mechanisms differ.) We find that for networks wide enough, the mixed models of $\pi(\Theta_B)$ and $\Theta_A$ have high accuracy and low loss value. We also observe that the network width remains a necessary condition for model combination to succeed.

In conclusion, our results reveal that \emph{the implied volume where the combination `works' is vastly larger than the line segment connecting $\pi(\Theta_B)$ and $\Theta_A$, the main interest of former linear mode connectivity research.}

\subsection{Searching for parts of the hyperrectangle that `do not work'}

All our our model combination efforts so far resulted in `working' models. We now aim to purposely construct points in the hypercube which correspond to model combinations we expect to have high loss values --- i.e., `do not work'. We intend to design convex combinations that disrupt the network functionality with having in mind either the inner structure of convolutional filter pairs, or their supposed interplay or covariance.

\subsubsection{Always choosing the smaller or the larger weight}

\begin{figure}
    \centering
    \subfloat[Loss]{%
        \includegraphics[trim=0cm 0cm 0cm 0cm, clip=true, width=0.5\textwidth]{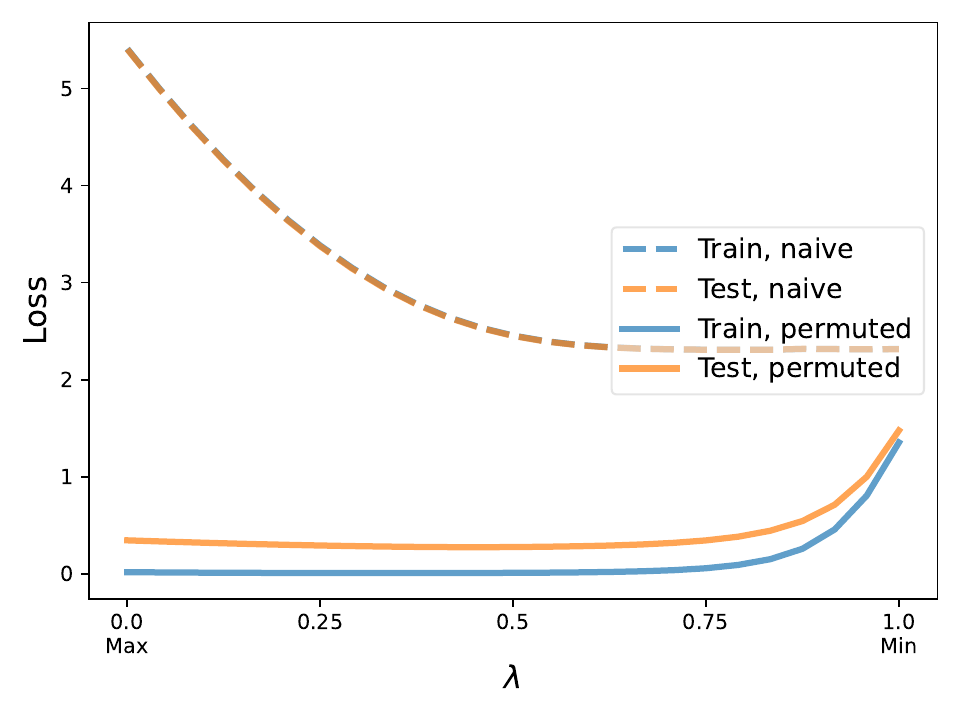}%
        \label{fig:minmax_abs_loss_resnet}%
        }%
    \hfill%
    \subfloat[Accuracy]{%
        \includegraphics[trim=0cm 0cm 0cm 0cm, clip=true, width=0.5\textwidth]{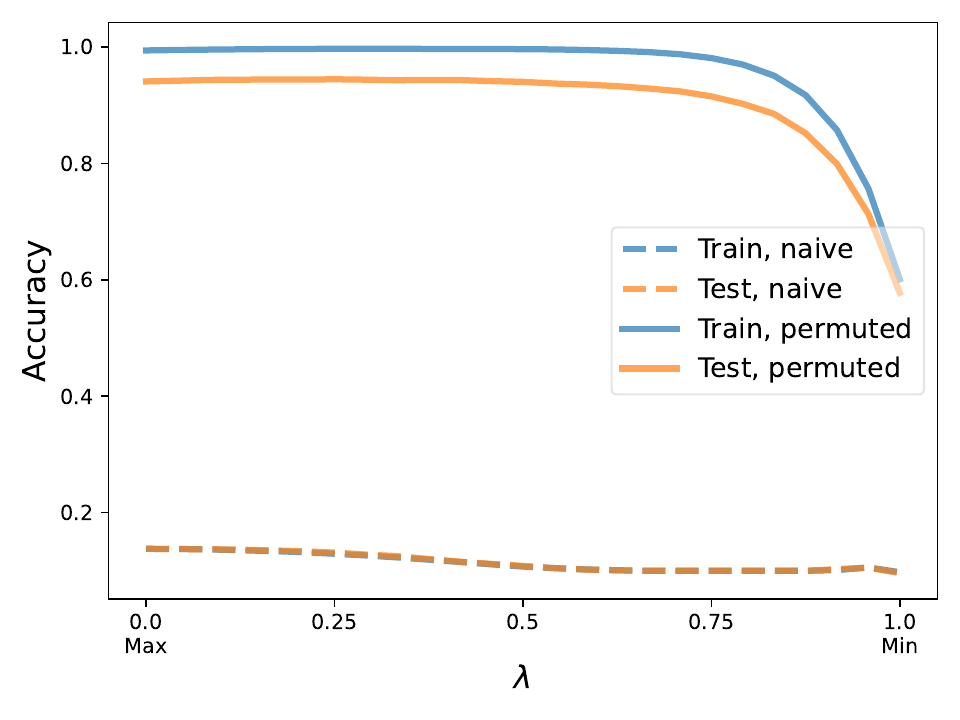}%
        \label{fig:minmax_acc_resnet}%
        }%
    \caption{Linear interpolation between the `Max' model and the `Min' model. The model was ResNet-20. Horizontal axes denote the $\lambda$ interpolation coefficient, vertical axes denote performance in terms of (a) loss and (b) accuracy.}
    \label{fig:minmax_abs_resnet}
\end{figure}

We find that when taking for each pair of weights $(w_a, w_b)$ the one for which the absolute value is smaller, this new `Min' model has low accuracy. Figure~\ref{fig:minmax_abs_resnet} shows linear interpolation between the `Min' model and the `Max' model, where for each weight pair we choose the one with the largest absolute value. 

This suits as a counterexample to the conjecture that \emph{any} point of the hypercube leads to a well-performing combination.
\section{Model combinations beyond the hyperrectangle}
\label{sec:combinations_beyond_the_hyperrectangle}

While the hyperrectangle spanned by $\Theta_A$ and $\pi(\Theta_B)$ provides a natural space of possibilities for model combinations, we do not suppose that the faces of this box serve as a hard boundary for the loss basin. Conversely, it is reasonable to anticipate that the low loss region expands beyond the extremal points. In this section, we test this hypothesis by exploring such combinations.

\subsection{Linear extrapolation}

\begin{figure}
    \centering
    \subfloat[Loss]{%
        \includegraphics[trim=0cm 0cm 0cm 0cm, clip=true, width=0.5\textwidth]{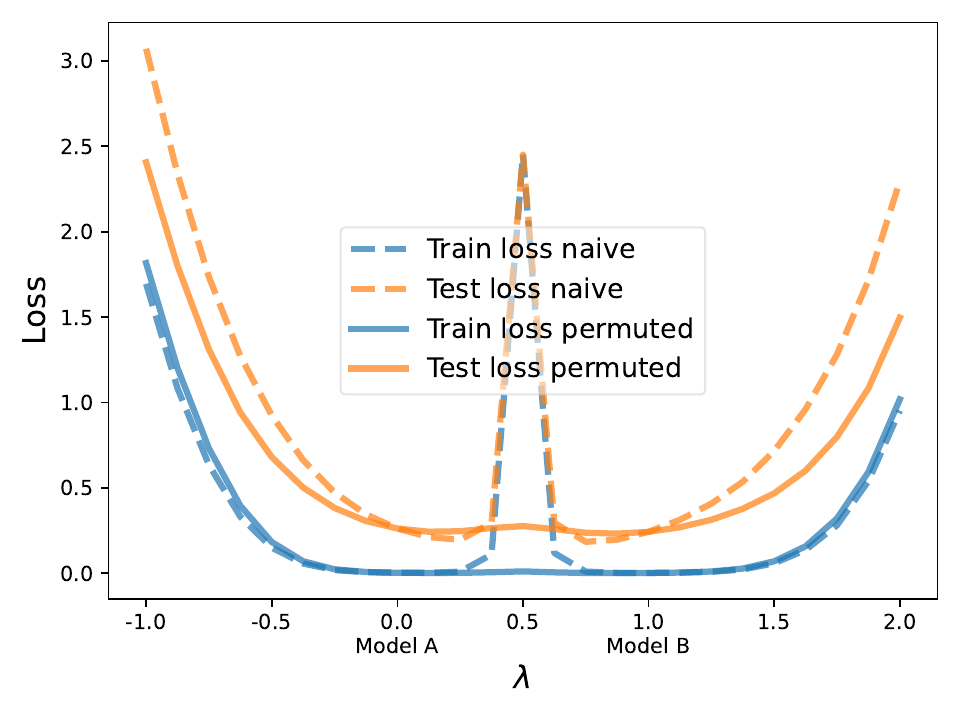}%
        \label{fig:linear_extrapolation_resnet_loss}%
        }%
    \hfill%
    \subfloat[Accuracy]{%
        \includegraphics[trim=0cm 0cm 0cm 0cm, clip=true, width=0.5\textwidth]{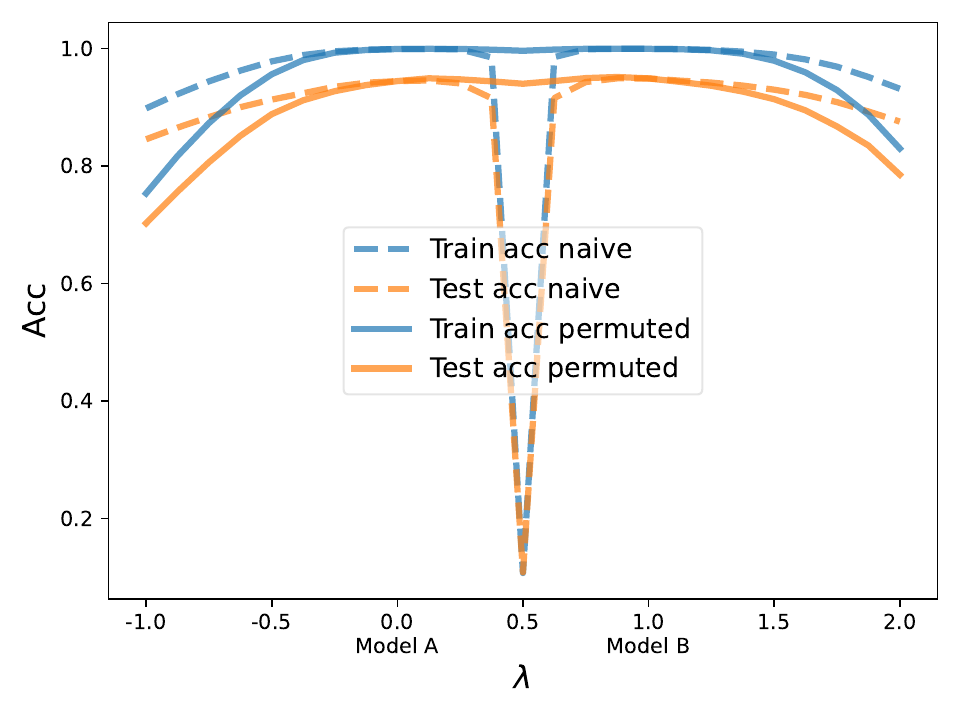}%
        \label{fig:linear_extrapolation_resnet_acc}%
        }%
    \caption{Linear extrapolation with models trained on ResNet-20. Horizontal axes denote the $\lambda$ interpolation coefficient, vertical axes denote performance in terms of (a) loss and (b) accuracy.}
    \label{fig:linear_extrapolation_resnet}
\end{figure}

First, as perhaps the simplest approach, we extend the range of the linear interpolation coefficient $\lambda \in \mathbb{R}$ from the unit interval to $[-1, 2]$ (note that the subinterval [0, 1] corresponds to the original linear interpolation between model A and model B). Figure~\ref{fig:linear_extrapolation_resnet} depicts the results for ResNet-20. We can observe that the extrapolated models perform well within certain segments of the extended range. Specifically, within the intervals of $\lambda \in [-0.25, 0]$ and $\lambda \in [1.0, 1.25]$, the train loss and accuracy of the resulting model combinations are virtually identical to the originals. On the test set, the performance starts to degrade slowly as we pass the boundaries of the unit interval, still resulting in considerable segments of well-generalizing models. We can also observe that with permuted models the performance degrades significantly slower than with the naïve counterparts. This suggests that aligned weight pairs result in a less disruptive perturbation regarding functionality.

\begin{figure}
    \centering
    \subfloat[Train loss]{%
        \includegraphics[trim=0cm 0cm 0cm 0cm, clip=true, width=0.5\textwidth]{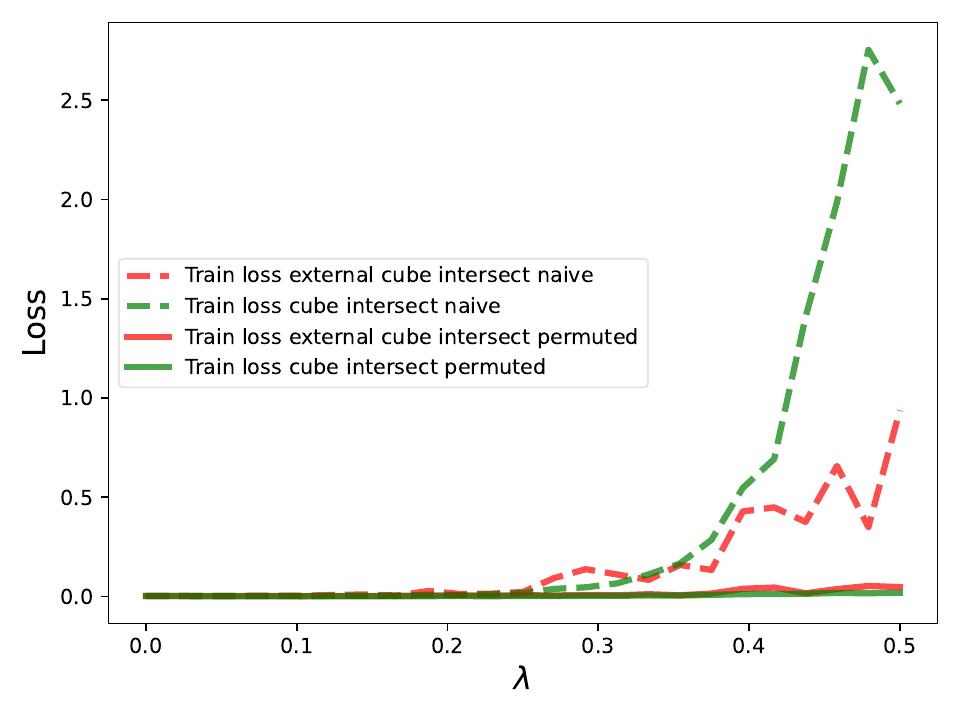}%
        \label{fig:multi_distr_extrapolation_resnet_loss}%
        }%
    \hfill%
    \subfloat[Test loss]{%
        \includegraphics[trim=0cm 0cm 0cm 0cm, clip=true, width=0.5\textwidth]{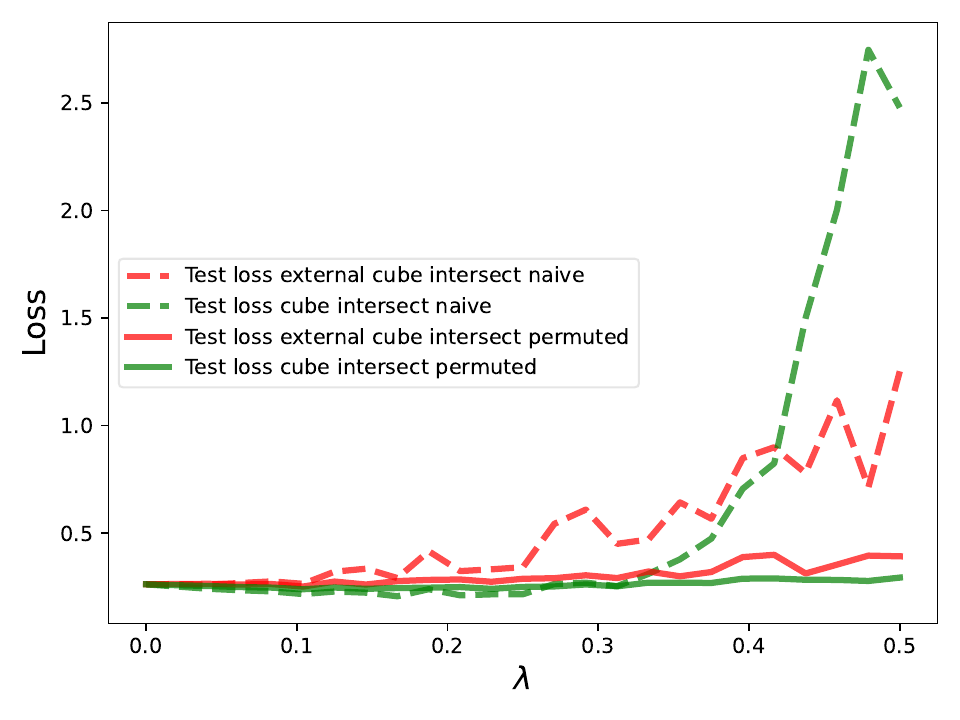}%
        \label{fig:multi_distr_extrapolation_resnet_acc}%
        }%
    \caption{Performance of ResNet-20 model combinations sampled uniformly from $\Theta_A$ centered hyperrectangles. Horizontal axes is the scaling parameter of the box, vertical axes represent performance in terms of loss.}
    \label{fig:multi_distr_extrapolation_resnet}
\end{figure}

\subsection{Sampling uniformly from $\Theta_A$ centered hyperrectangles}

\label{sec:shifted_uniform}

We now turn our attention to element-wise model combinations which extrapolate. As an instance of such an experiment, we shift the uniform sampling from the hyperrectangle spanned by $\Theta_A$ and $\pi(\Theta_B)$ to be centered at $\Theta_A$. This corresponds to model combinations where negative terms are also allowed, i.e., we choose our combining coefficients from $[-s, s]^d$ for a given $s \in \mathbb{R}$. This way, we obtain a box that corresponds to perturbed versions of $\Theta_A$ obtained by applying an additive uniform noise scaled according to the resulting weight differences of the alignment process. (Informally, if a weight matched to a weight with a similar value, that corresponds to a small edge of the box, and thus, results in a small uniform perturbation of that weight; the case of large differences plays out analogously.) In this way, we connect the magnitude of the applied noise to the `degree of determination' of weight values resulting from the matching procedure. Figure~\ref{fig:multi_distr_extrapolation_resnet} depicts the results for various $s \in [0, 0.5]$. We observe that the model combinations of aligned models work well for $s \in [0, 0.25]$.

\subsection{Transitivity of linear mode connectivity}

\begin{figure}[h]
    \centering
    \subfloat[Interpolating $\pi_B(\Theta_B)$ and $\pi_C(\Theta_C)$]{%
        \centering
        \includegraphics[trim=0cm 0cm 0cm 0cm, clip=true, width=0.49\linewidth]{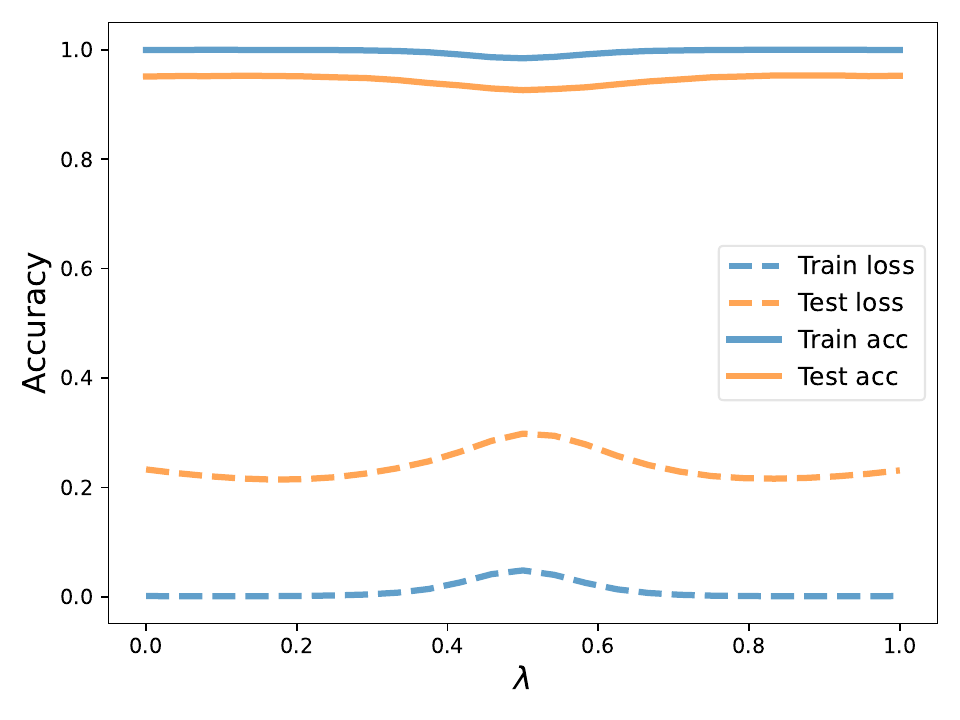}%
        \label{fig:transitivity_pair_resnet}
    }%
    \hfill%
    \subfloat[Accuracy heatmap]{%
        \includegraphics[trim=0cm 0cm 0cm 0cm, clip=true, width=0.49\linewidth]{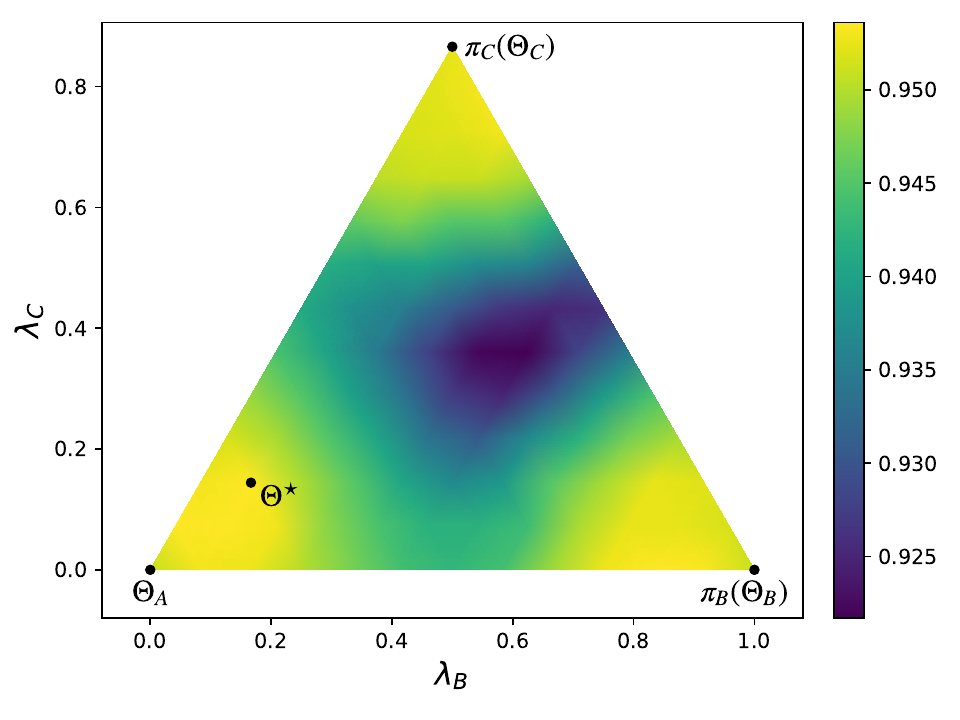}%
        \label{fig:transitivity_triangle_resnet}%
        }%
    \caption{(a) Interpolating between two models $\pi_B(\Theta_B)$ and $\pi_C(\Theta_C)$, where the permutations $\pi_B$ and $\pi_C$ re-basin models $\Theta_B$ and $\Theta_C$ to the basin of a common model $\Theta_A$. (b) Test accuracy heatmap for linear combinations of three models: $\Theta_A$, $\pi_B(\Theta_B)$, and $\pi_C(\Theta_C)$. For visual convenience, the coefficients $\lambda_B$ and $\lambda_C$ are transformed to form an equilateral triangle. $\Theta^{\star}$ denote the best performing model. The model is ResNet-20 with a width multiplier of 40 to highlight an example where the best performing model is an interior point.}
\end{figure}

The question naturally arises whether two models $\pi_B(\Theta_B)$ and $\pi_C(\Theta_C)$ re-basined to a third model $\Theta_A$ are also linearly interpolable. We answer this question affirmatively. Figure~\ref{fig:transitivity_pair_resnet} depicts such configurations, and we can observe high accuracy and low loss values for these combinations.

\subsection{Combining three models}

We also investigate the convex combinations of the three models. Figure~\ref{fig:transitivity_triangle_resnet} depict such convex combinations written up in the form $\Theta_A + \lambda_B \cdot ( \pi_B(\Theta_B)-\Theta_A) + \lambda_C \cdot (\pi_C(\Theta_C)-\Theta_A)$, where $\lambda_B, \lambda_C \in \mathbb{R}$ are the interpolation coefficients controlling the extent to which we step from $\Theta_A$ in the direction of $\pi_B(\Theta_B)$ or $\pi_C(\Theta_C)$. The figure shows that the test accuracy of even the worst performing model is above $0.91$ still which we can consider a well-performing model.

There are more observations to be made in Figure~\ref{fig:transitivity_triangle_resnet}. First, combined models can surpass the performance of the originals; second, in the depicted example the best-performing model is at an interior point of the depicted triangle. While similar observations of better-performing interpolated models have been made in \citet[see Figure 5]{ainsworth2022rebasin}, only in the case when they utilize a split data training (merging two models trained with two disjoint datasets on CIFAR-10), or in the case of a much simpler dataset of MNIST. In our case, the training dataset is the same for all source models, and the CIFAR-10 classification task is much more complex than that of MNIST. We observed the same behavior for ResNet-20 models of different width multipliers ranging from 10 to 40. However, in many cases, the best performing model was on an edge of the triangle. Overall, the above observations suggest that even slight model differences --- originating only from the different weight initializations --- can be exploited to obtain better performing combined models, and it is worth pursuing combining multiple models.

Also note, that in this case, we interpolate using only two real parameters. This naturally leads to the question of whether more sophisticated (e.g., well-chosen element-wise) model combinations could result in even higher performance. In light of the experiments presented in this paper, we find this an intriguing research direction. However, this question is beyond the scope of this article, and we leave it for future work.

\section{Functional and weight dissimilarity of model combinations}
\label{sec:func_and_weight_similiarity}

In the abundance of well-performing model combinations, it is reasonable to ask whether these combinations are vacuous in the sense that the original models $\Theta_A$ and $\pi(\Theta_B)$ could be nearly identical, rendering the space of combinations effectively empty. While the examples above of model combinations surpassing the performance of the originals already suggest the contrary, we are not aware of any work in the literature that looks `under the hood' to highlight such differences in a more detailed manner. In this section, we conduct several experiments to demonstrate that there are indeed significant functional differences both between the original models and their combinations.

\subsection{Functional difference of model combinations}

\emph{Do the original models or their combinations output the same labels?} --- First, we investigate network-level functional similarity by comparing the predictions of different model combinations. We use the predicted labels as the ground of comparison as these serve as an easily interpretable aggregation of model functionality. More precisely, to test a model combination for its similarity to the original models $\Theta_A$ and $\Theta_B$, we take the test set of the CIFAR-10 dataset and count how many of the data points fall into each of the following four categories: the label prediction of the output matches the prediction of $\Theta_A$ only, $\Theta_B$ only, neither, or both. In each category, the lighter and darker shade depict whether the prediction of the model matches the true label (`correct') or not (`wrong'). 

\begin{figure}
\centering
    \subfloat[Bernoulli]{%
        \includegraphics[trim=0cm 0cm 0cm 0cm, clip=true, width=0.49\textwidth]{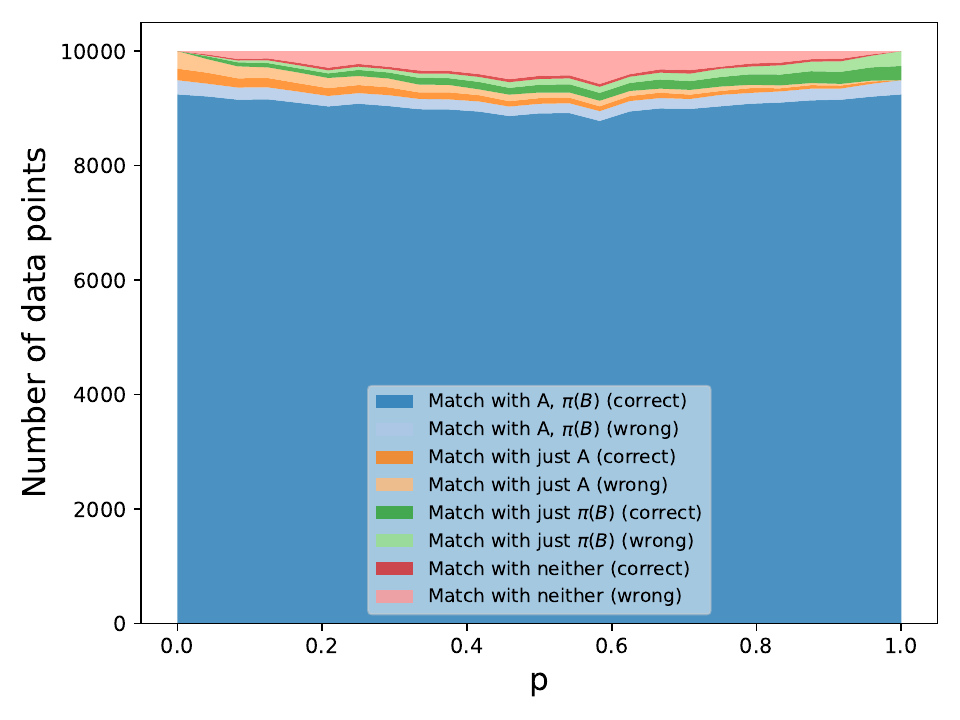}%
        \label{fig:stack_plot_bernoulli_resnet}%
        }%
    \subfloat[Cube intersect]{%
        \includegraphics[trim=0cm 0cm 0cm 0cm, clip=true, width=0.49\textwidth]{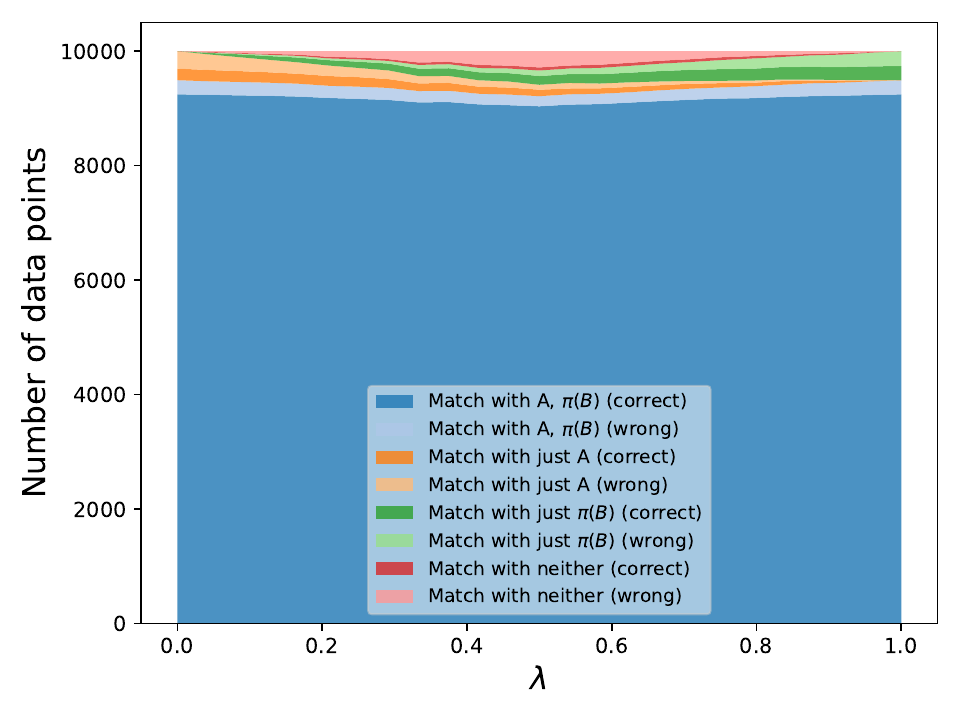}%
        \label{fig:stack_plot_cube_intersect_resnet}%
        }%
        
    \subfloat[Plane intersect]{%
        \includegraphics[trim=0cm 0cm 0cm 0cm, clip=true, width=0.49\textwidth]{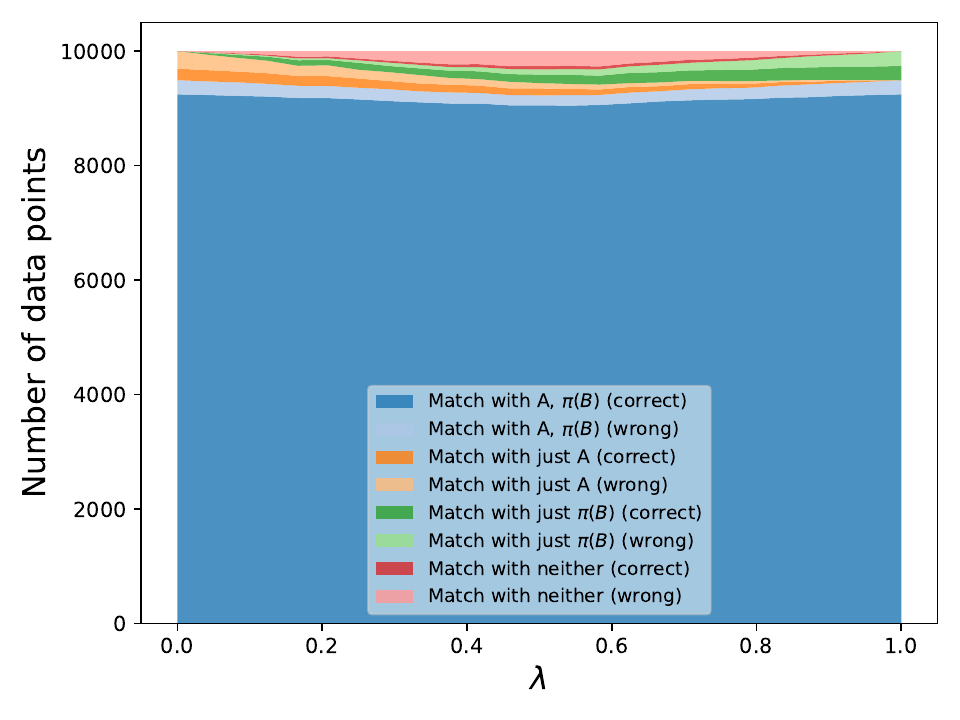}%
        \label{fig:stack_plot_plane_intersect_resnet}%
        }%
    \subfloat[Uniform]{%
        \includegraphics[trim=0cm 0cm 0cm 0cm, clip=true, width=0.49\textwidth]{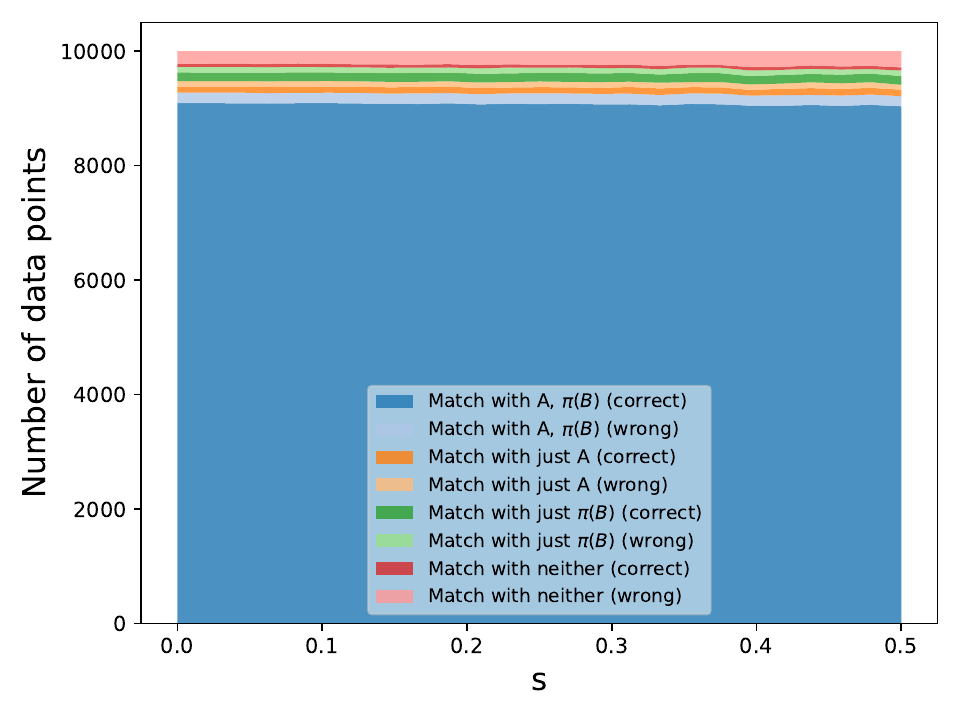}%
        \label{fig:stack_plot_uniform_resnet}%
        }%
    \caption{Network level functional comparison of ResNet-20 model combinations. 
    The stacked plots illustrate the distribution of the 10,000 CIFAR-10 test datapoints for different model combinations across four categories: the predicted labels match with Model A only, Model B only, neither, or both.}
    \label{fig:stack_plots_resnet}
\end{figure}

Figure~\ref{fig:stack_plots_resnet} depicts the results for different model combinations. For each category we also represent whether the predicted label matches the true label. We can observe that in the case of linear interpolation, Bernoulli, cube intersect, and plane intersect combinations of the two endpoints (corresponding precisely to $\Theta_A$ and $\pi(\Theta_B)$) \emph{show significant functional difference}, i.e., there is a substantial fraction of the test set for which they give differing labels. By tracking the variations along the horizontal axis in these figures, we can observe that these differences change gradually when interpolating between the endpoints. Obviously, we see a different picture for uniform combinations; in this case, both endpoints correspond to an already mixed model.

\subsection{Edge lengths of the hyperrectangles}

\begin{figure}
    \centering
    \includegraphics[trim=0cm 0cm 0cm 0cm, clip=true, width=0.5\textwidth]{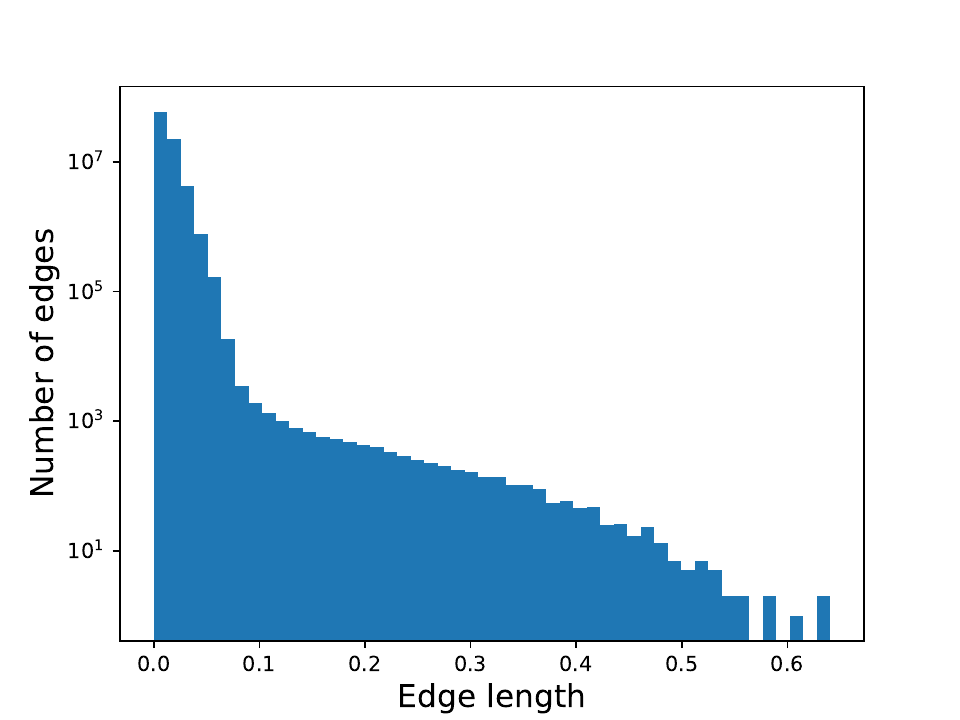}%
    \caption{Edge lengths of a hyperrectangle. Network is Tiny-10, width multiplier is 16.}
    \label{fig:edge_lengths}
\end{figure}

\emph{How much the weights of the two original models differ?} --- We now consider the element-wise weight differences of aligned model pairs $\Theta_A$ and $\pi(\Theta_B)$. Figure~\ref{fig:edge_lengths} shows the distribution of edge lengths (i.e., absolute differences of aligned weight pairs) for the hyperrectangles spanned by $\Theta_A$ and $\pi(\Theta_B)$. On one hand, we can observe that there are a diverse variety of lengths, and a \emph{significant portion of them differ from zero}. On the other hand, there is a large portion of near-zero lengths. We attribute this fact to the general sparsity property of deep neural networks that implies a large portion of low weight values. (Also, we used weight decay in our training procedure which might further reinforce this feature.)

\subsection{Robustness to alignment perturbations}
\label{sec:perturb_the_alignment}

\emph{Are all the filter pairings important?} --- To measure the robustness (or conversely, the vulnerability) of filter pair matchings regarding model performance, we deliberately corrupt parts of the alignment and measure how much it affects the accuracy of the combined model. First, for a given layer we order the filter pairs based on their activation correlation in a decreasing order. Then, we re-match randomly the $k$ lowest scoring pairs by taking a derangement $\pi_r$ (a random permutation where no element remains in its original position) on the $k$ filters. A specific property of this perturbation is that it maintains the overall magnitude and the distribution of activations of a layer while disturbing the filter pairings for the interpolation. To measure combination performance we interpolate between the resulting $\pi_r(\pi(\Theta_B))$ and $\Theta_A$ for a given layer and report accuracy barriers. (Note that permutations of a single model preserve perfect functional similarity as the subsequent layer's input ordering is adjusted appropriately when re-basing. However, when we combine the perturbed model with another one, it is appropriate to measure how the perturbation affects model combination.)

\begin{figure}
        \begin{subfigure}[t]{0.25\textwidth}
            \includegraphics[width=\textwidth]{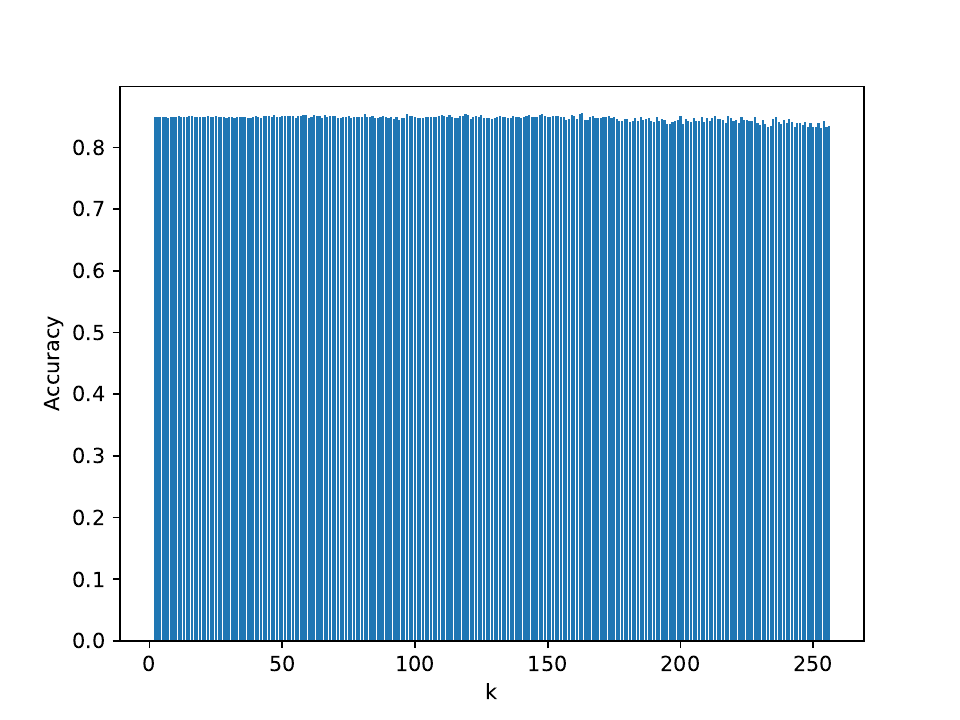}
            \label{fig:corrupt1}
        \end{subfigure}%
        \begin{subfigure}[t]{0.25\textwidth}
            \includegraphics[width=\textwidth]{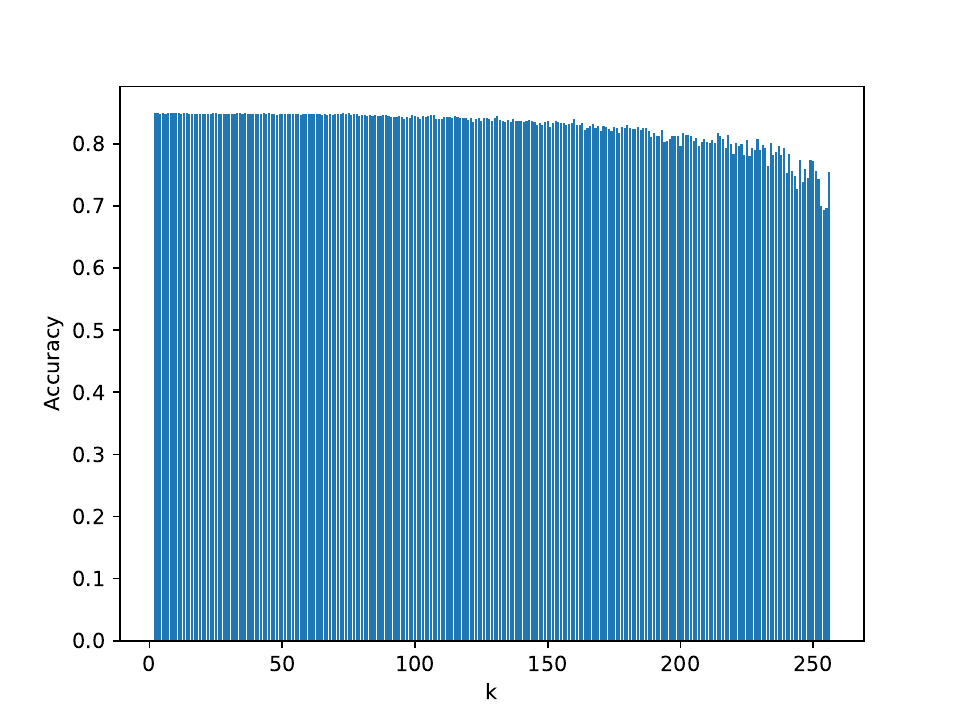}
            \label{fig:corrupt2}
        \end{subfigure}%
        \begin{subfigure}[t]{0.25\textwidth}
            \includegraphics[width=\textwidth]{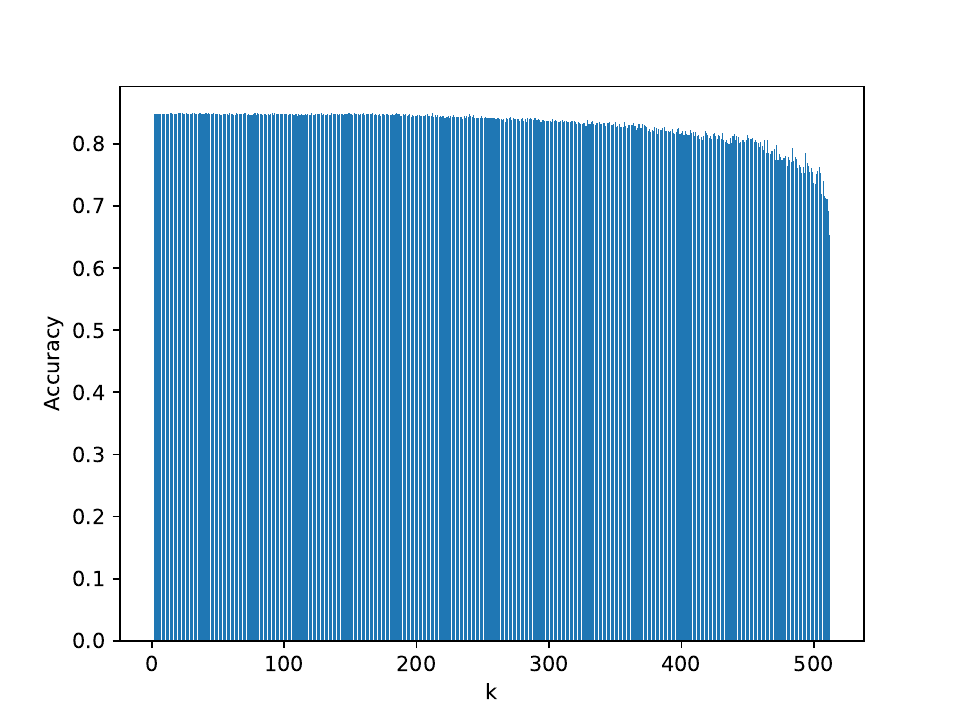}
            \label{fig:corrupt3}
        \end{subfigure}%
        \begin{subfigure}[t]{0.25\textwidth}
            \includegraphics[width=\textwidth]{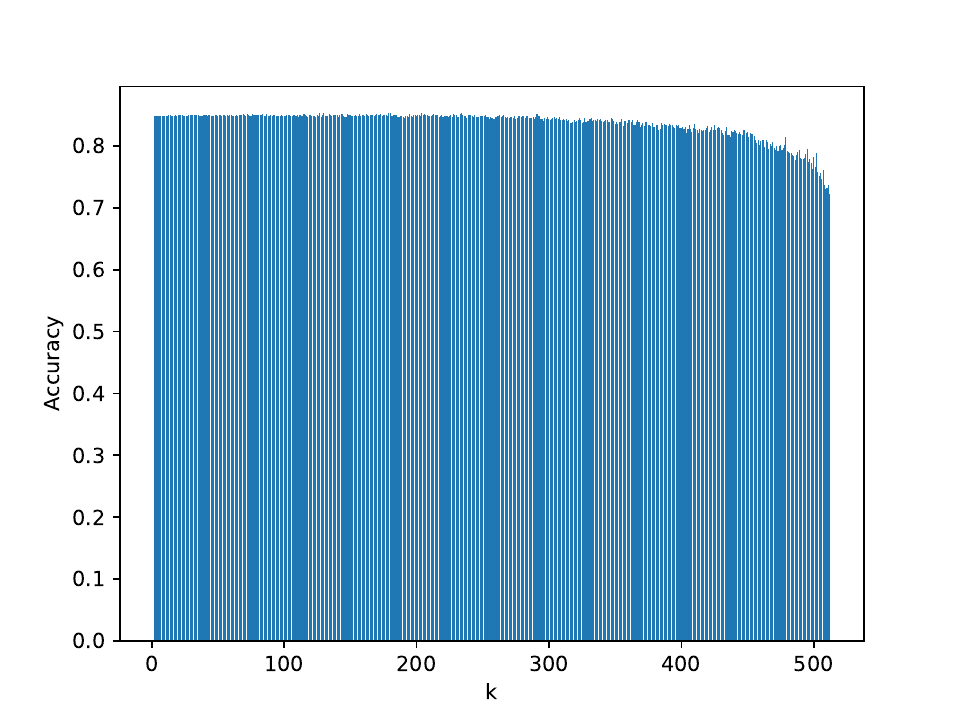}
            \label{fig:corrupt4}
        \end{subfigure}
    
        \begin{subfigure}[t]{0.25\textwidth}
            \includegraphics[width=\textwidth]{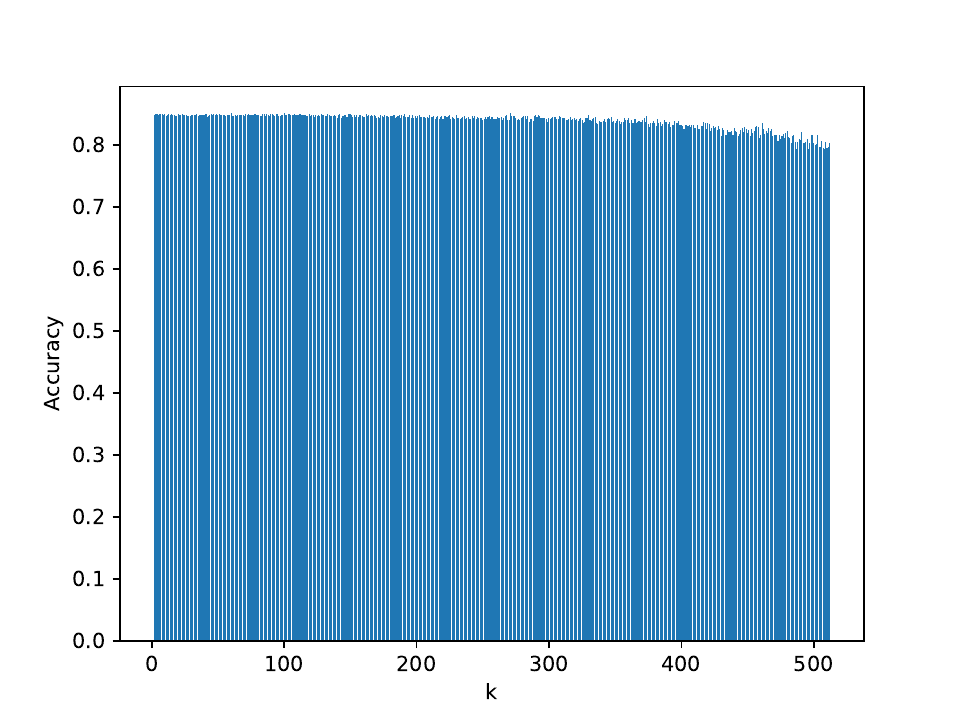}
            \label{fig:corrupt5}
        \end{subfigure}%
        \begin{subfigure}[t]{0.25\textwidth}
            \includegraphics[width=\textwidth]{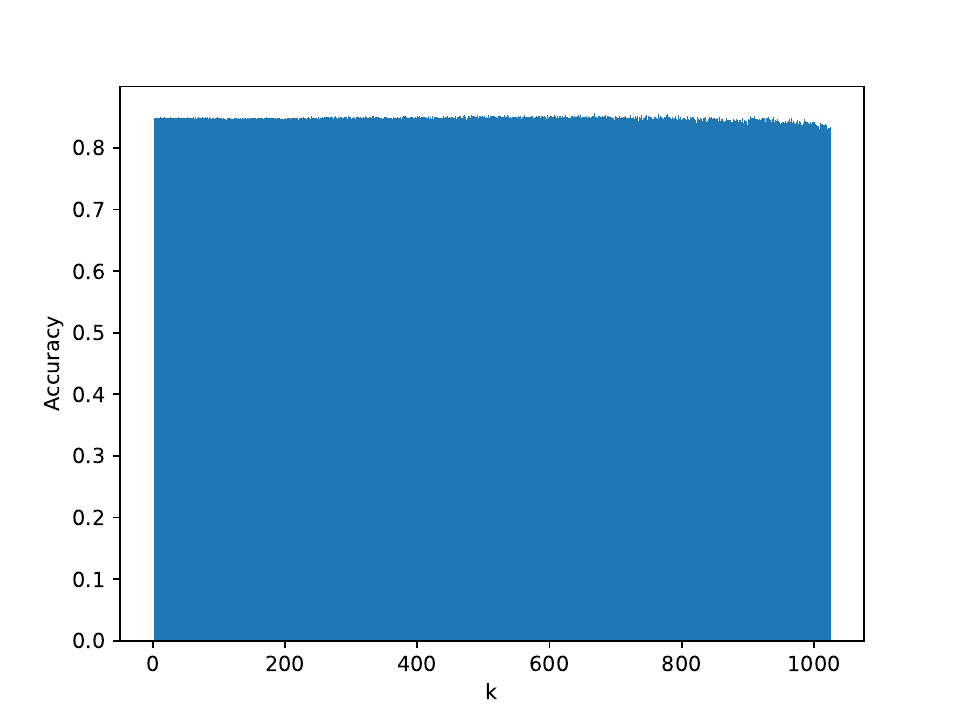}
            \label{fig:corrupt6}
        \end{subfigure}%
        \begin{subfigure}[t]{0.25\textwidth}
            \includegraphics[width=\textwidth]{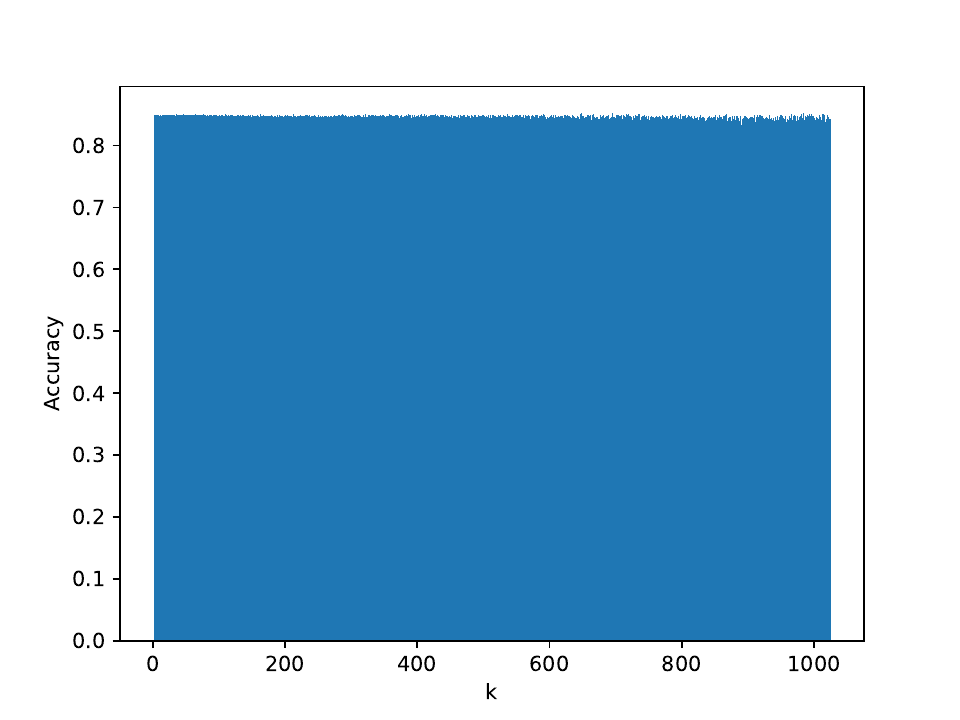}
            \label{fig:corrupt7}
        \end{subfigure}%
        \begin{subfigure}[t]{0.25\textwidth}
            \includegraphics[width=\textwidth]{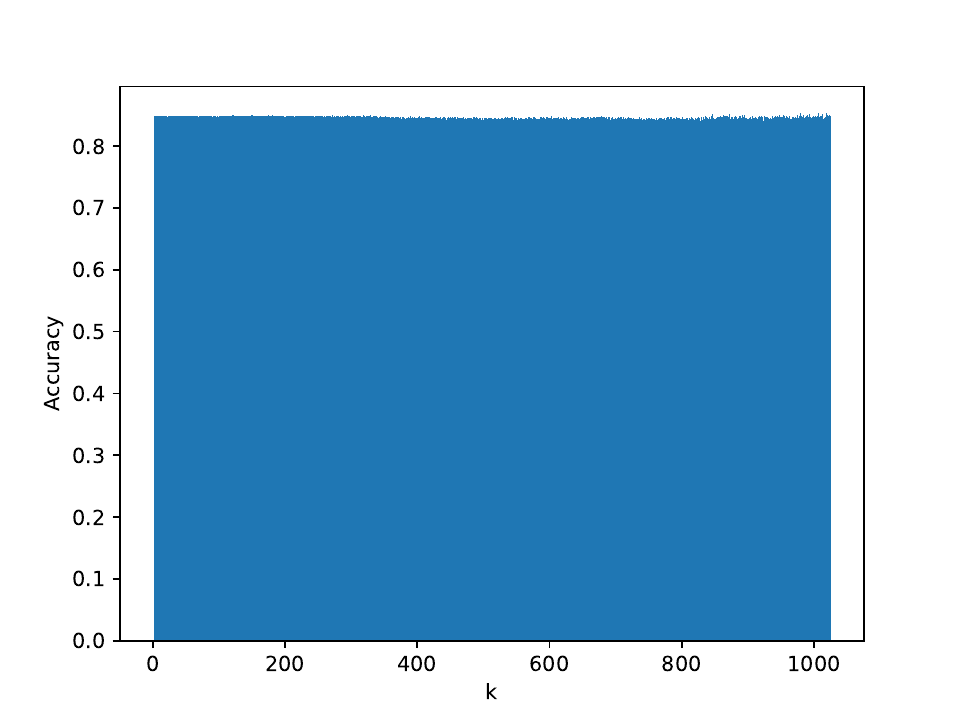}
            \label{fig:corrupt8}
        \end{subfigure}
        
        \caption{Measuring filter pair importance by randomly re-matching the lowest scoring $k$ pairs. Horizontal axes denotes the number of re-matched filters, vertical axes denote the loss barrier interpolating between the resulting models.}
        \label{fig:corrupt_pairs}
\end{figure}

Figure~\ref{fig:corrupt_pairs} depicts the loss and accuracy barrier for resulting model pairs with values of $k$ ranging between 2 and the number of filters $N$ in the given layer. \emph{We observe strong robustness to the alignment perturbations.} For each layer, a large portion of the alignment can be randomly interchanged. Remarkably, for Layers 1, 6, 7, and 8, even a complete derangement ($k=N$) leaves the combined networks well-performing. For the other layers, we can observe mild degradation of performance for up to half of the filters, where \emph{the deterioration starts to accelerate}.

With a derangement $\pi_r(\Theta_B)$, we replace an aligned filter with a randomly chosen other one. When interpolating, the filters of $\Theta_A$ thus are exposed to a random noise distributed as the corresponding filter weights of $\Theta_B$. (We hypothesize that this distribution closely mirrors the distribution of filter weights in $\Theta_A$ itself.) We attribute the above remarkable observations of robustness to the fact that this distribution has a large probability mass on near zero weights, or on weights whose difference is smaller than the supposed extent of noise resilience of filter functionality.

In this experiment, we ordered the filter pairs according to activation correlations. We speculate that filters that are more crucial for the overall network functionality might appear in both networks more unambiguously, can be identified in the process of re-basing more clearly, and as a pair, they exhibit larger activation correlations. The disruption of these important filters is what we might observe on the right-hand side of these figures. One might consider including further filter importance measures and model pruning in an analysis, but such explorations extend beyond the scope of the present paper, and we leave it for future work.

\section{Conclusion}
\label{sec:conclusion}

We showcased numerous model combination methods, indicating that the loss basin studied by previous LMC research is much broader than formerly identified. This demonstrates that LMC is a special case of a more general model combinability phenomenon. We investigated model combinations in terms of performance. We examined weight and functional similarities on several granularities and identified a general robustness and noise resilience property.

Model combinations in a more general sense --- going back to, e.g., classical ensemble learning --- were always an important aspect of practical and theoretical machine learning. Deep learning models exhibit immense complexity, yet, model combination in the weight space is developing as a promising direction for knowledge aggregation and extending generalization capabilities. Uncovering and exploring novel ways of model combinations, and extending work on the symmetries and regularities in deep learning models help taming such complexities and foster developing methods that exploit model combinability.

\section*{Acknowledgements}

This work was supported by the Ministry of Innovation and Technology NRDI Office within the framework of the Artificial Intelligence National Laboratory (RRF-2.3.1-21-2022-00004). A. Cs. was partly supported by the project TKP2021-NKTA-62 financed by the National Research, Development and Innovation Fund of the Ministry for Innovation and Technology, Hungary.

\bibliographystyle{elsarticle-harv}
\bibliography{references}

\begin{thebibliography}{22}
\expandafter\ifx\csname natexlab\endcsname\relax\def\natexlab#1{#1}\fi
\providecommand{\url}[1]{\texttt{#1}}
\providecommand{\href}[2]{#2}
\providecommand{\path}[1]{#1}
\providecommand{\DOIprefix}{doi:}
\providecommand{\ArXivprefix}{arXiv:}
\providecommand{\URLprefix}{URL: }
\providecommand{\Pubmedprefix}{pmid:}
\providecommand{\doi}[1]{\href{http://dx.doi.org/#1}{\path{#1}}}
\providecommand{\Pubmed}[1]{\href{pmid:#1}{\path{#1}}}
\providecommand{\bibinfo}[2]{#2}
\ifx\xfnm\relax \def\xfnm[#1]{\unskip,\space#1}\fi
\bibitem[{Ainsworth et~al.(2023)Ainsworth, Hayase and
  Srinivasa}]{ainsworth2022rebasin}
\bibinfo{author}{Ainsworth, S.}, \bibinfo{author}{Hayase, J.},
  \bibinfo{author}{Srinivasa, S.}, \bibinfo{year}{2023}.
\newblock \bibinfo{title}{Git re-basin: Merging models modulo permutation
  symmetries}, in: \bibinfo{booktitle}{The Eleventh International Conference on
  Learning Representations}.
\newblock \URLprefix \url{https://openreview.net/forum?id=CQsmMYmlP5T}.
\bibitem[{Ba et~al.(2016)Ba, Kiros and Hinton}]{ba2016layer}
\bibinfo{author}{Ba, J.L.}, \bibinfo{author}{Kiros, J.R.},
  \bibinfo{author}{Hinton, G.E.}, \bibinfo{year}{2016}.
\newblock \bibinfo{title}{Layer normalization}.
\newblock \bibinfo{journal}{arXiv preprint arXiv:1607.06450} .
\bibitem[{Bansal et~al.(2021)Bansal, Nakkiran and Barak}]{bansal2021revisiting}
\bibinfo{author}{Bansal, Y.}, \bibinfo{author}{Nakkiran, P.},
  \bibinfo{author}{Barak, B.}, \bibinfo{year}{2021}.
\newblock \bibinfo{title}{Revisiting model stitching to compare neural
  representations}.
\newblock \bibinfo{journal}{Advances in Neural Information Processing Systems}
  \bibinfo{volume}{35}.
\bibitem[{Bertsekas(1998)}]{bertsekas1998network}
\bibinfo{author}{Bertsekas, D.}, \bibinfo{year}{1998}.
\newblock \bibinfo{title}{Network Optimization: Continuous and Discrete
  Methods}.
\newblock Athena scientific optimization and computation series,
  \bibinfo{publisher}{Athena Scientific}.
\newblock \URLprefix \url{https://books.google.com/books?id=afYYAQAAIAAJ}.
\bibitem[{Csisz{\'a}rik et~al.(2021)Csisz{\'a}rik, K{\H{o}}r{\"o}si-Szab{\'o},
  Matszangosz, Papp and Varga}]{csiszarik2021similarity}
\bibinfo{author}{Csisz{\'a}rik, A.},
  \bibinfo{author}{K{\H{o}}r{\"o}si-Szab{\'o}, P.},
  \bibinfo{author}{Matszangosz, {\'A}.}, \bibinfo{author}{Papp, G.},
  \bibinfo{author}{Varga, D.}, \bibinfo{year}{2021}.
\newblock \bibinfo{title}{Similarity and matching of neural network
  representations}.
\newblock \bibinfo{journal}{Advances in Neural Information Processing Systems}
  \bibinfo{volume}{34}, \bibinfo{pages}{5656--5668}.
\bibitem[{Dettmers et~al.(2023)Dettmers, Pagnoni, Holtzman and
  Zettlemoyer}]{dettmers2023qlora}
\bibinfo{author}{Dettmers, T.}, \bibinfo{author}{Pagnoni, A.},
  \bibinfo{author}{Holtzman, A.}, \bibinfo{author}{Zettlemoyer, L.},
  \bibinfo{year}{2023}.
\newblock \bibinfo{title}{Qlora: Efficient finetuning of quantized llms}.
\newblock \href{http://arxiv.org/abs/2305.14314}{{\tt arXiv:2305.14314}}.
\bibitem[{Draxler et~al.(2018)Draxler, Veschgini, Salmhofer and
  Hamprecht}]{draxler2018essentially}
\bibinfo{author}{Draxler, F.}, \bibinfo{author}{Veschgini, K.},
  \bibinfo{author}{Salmhofer, M.}, \bibinfo{author}{Hamprecht, F.},
  \bibinfo{year}{2018}.
\newblock \bibinfo{title}{Essentially no barriers in neural network energy
  landscape}, in: \bibinfo{booktitle}{International conference on machine
  learning}, \bibinfo{organization}{PMLR}. pp. \bibinfo{pages}{1309--1318}.
\bibitem[{Entezari et~al.(2022)Entezari, Sedghi, Saukh and
  Neyshabur}]{entezari2022permutation}
\bibinfo{author}{Entezari, R.}, \bibinfo{author}{Sedghi, H.},
  \bibinfo{author}{Saukh, O.}, \bibinfo{author}{Neyshabur, B.},
  \bibinfo{year}{2022}.
\newblock \bibinfo{title}{The role of permutation invariance in linear mode
  connectivity of neural networks}, in: \bibinfo{booktitle}{International
  Conference on Learning Representations}.
\newblock \URLprefix \url{https://openreview.net/forum?id=dNigytemkL}.
\bibitem[{Frankle et~al.(2020)Frankle, Dziugaite, Roy and
  Carbin}]{frankle2020linear}
\bibinfo{author}{Frankle, J.}, \bibinfo{author}{Dziugaite, G.K.},
  \bibinfo{author}{Roy, D.}, \bibinfo{author}{Carbin, M.},
  \bibinfo{year}{2020}.
\newblock \bibinfo{title}{Linear mode connectivity and the lottery ticket
  hypothesis}, in: \bibinfo{booktitle}{International Conference on Machine
  Learning}, \bibinfo{organization}{PMLR}. pp. \bibinfo{pages}{3259--3269}.
\bibitem[{Garipov et~al.(2018)Garipov, Izmailov, Podoprikhin, Vetrov and
  Wilson}]{garipov2018loss}
\bibinfo{author}{Garipov, T.}, \bibinfo{author}{Izmailov, P.},
  \bibinfo{author}{Podoprikhin, D.}, \bibinfo{author}{Vetrov, D.P.},
  \bibinfo{author}{Wilson, A.G.}, \bibinfo{year}{2018}.
\newblock \bibinfo{title}{Loss surfaces, mode connectivity, and fast ensembling
  of dnns}.
\newblock \bibinfo{journal}{Advances in neural information processing systems}
  \bibinfo{volume}{31}.
\bibitem[{He et~al.(2016)He, Zhang, Ren and Sun}]{he2016deep}
\bibinfo{author}{He, K.}, \bibinfo{author}{Zhang, X.}, \bibinfo{author}{Ren,
  S.}, \bibinfo{author}{Sun, J.}, \bibinfo{year}{2016}.
\newblock \bibinfo{title}{Deep residual learning for image recognition}, in:
  \bibinfo{booktitle}{Proceedings of the IEEE conference on computer vision and
  pattern recognition}, pp. \bibinfo{pages}{770--778}.
\bibitem[{Hu et~al.(2022)Hu, yelong shen, Wallis, Allen-Zhu, Li, Wang, Wang and
  Chen}]{hu2022lora}
\bibinfo{author}{Hu, E.J.}, \bibinfo{author}{yelong shen},
  \bibinfo{author}{Wallis, P.}, \bibinfo{author}{Allen-Zhu, Z.},
  \bibinfo{author}{Li, Y.}, \bibinfo{author}{Wang, S.}, \bibinfo{author}{Wang,
  L.}, \bibinfo{author}{Chen, W.}, \bibinfo{year}{2022}.
\newblock \bibinfo{title}{Lo{RA}: Low-rank adaptation of large language
  models}, in: \bibinfo{booktitle}{International Conference on Learning
  Representations}.
\newblock \URLprefix \url{https://openreview.net/forum?id=nZeVKeeFYf9}.
\bibitem[{Ioffe and Szegedy(2015)}]{ioffe2015batch}
\bibinfo{author}{Ioffe, S.}, \bibinfo{author}{Szegedy, C.},
  \bibinfo{year}{2015}.
\newblock \bibinfo{title}{Batch normalization: Accelerating deep network
  training by reducing internal covariate shift}, in:
  \bibinfo{booktitle}{International conference on machine learning},
  \bibinfo{organization}{pmlr}. pp. \bibinfo{pages}{448--456}.
\bibitem[{Jordan et~al.(2022)Jordan, Sedghi, Saukh, Entezari and
  Neyshabur}]{jordan2022repair}
\bibinfo{author}{Jordan, K.}, \bibinfo{author}{Sedghi, H.},
  \bibinfo{author}{Saukh, O.}, \bibinfo{author}{Entezari, R.},
  \bibinfo{author}{Neyshabur, B.}, \bibinfo{year}{2022}.
\newblock \bibinfo{title}{Repair: Renormalizing permuted activations for
  interpolation repair}.
\newblock \URLprefix \url{https://arxiv.org/abs/2211.08403},
  \DOIprefix\doi{10.48550/ARXIV.2211.08403}.
\bibitem[{Kornblith et~al.(2019)Kornblith, Norouzi, Lee and
  Hinton}]{kornblith2019similarity}
\bibinfo{author}{Kornblith, S.}, \bibinfo{author}{Norouzi, M.},
  \bibinfo{author}{Lee, H.}, \bibinfo{author}{Hinton, G.},
  \bibinfo{year}{2019}.
\newblock \bibinfo{title}{Similarity of neural network representations
  revisited}, in: \bibinfo{booktitle}{International Conference on Machine
  Learning}, \bibinfo{organization}{PMLR}. pp. \bibinfo{pages}{3519--3529}.
\bibitem[{Krizhevsky and Hinton(2009)}]{krizhevsky2009learning}
\bibinfo{author}{Krizhevsky, A.}, \bibinfo{author}{Hinton, G.},
  \bibinfo{year}{2009}.
\newblock \bibinfo{title}{Learning multiple layers of features from tiny
  images}.
\newblock \bibinfo{type}{Technical Report}. University of Toronto.
  \bibinfo{address}{Toronto, Ontario}.
\bibitem[{Lenc and Vedaldi(2019)}]{lenc_vedaldi_19}
\bibinfo{author}{Lenc, K.}, \bibinfo{author}{Vedaldi, A.},
  \bibinfo{year}{2019}.
\newblock \bibinfo{title}{Understanding image representations by measuring
  their equivariance and equivalence}.
\newblock \bibinfo{journal}{International Journal of Computer Vision}
  \bibinfo{volume}{127}.
\newblock \DOIprefix\doi{10.1007/s11263-018-1098-y}.
\bibitem[{Matena and Raffel(2022)}]{matena2022merging}
\bibinfo{author}{Matena, M.S.}, \bibinfo{author}{Raffel, C.A.},
  \bibinfo{year}{2022}.
\newblock \bibinfo{title}{Merging models with fisher-weighted averaging}.
\newblock \bibinfo{journal}{Advances in Neural Information Processing Systems}
  \bibinfo{volume}{35}, \bibinfo{pages}{17703--17716}.
\bibitem[{McMahan et~al.(2017)McMahan, Moore, Ramage, Hampson and
  y~Arcas}]{mcmahan2017communication}
\bibinfo{author}{McMahan, B.}, \bibinfo{author}{Moore, E.},
  \bibinfo{author}{Ramage, D.}, \bibinfo{author}{Hampson, S.},
  \bibinfo{author}{y~Arcas, B.A.}, \bibinfo{year}{2017}.
\newblock \bibinfo{title}{Communication-efficient learning of deep networks
  from decentralized data}, in: \bibinfo{booktitle}{Artificial intelligence and
  statistics}, \bibinfo{organization}{PMLR}. pp. \bibinfo{pages}{1273--1282}.
\bibitem[{Meng et~al.(2022)Meng, Bau, Andonian and Belinkov}]{meng22_loced}
\bibinfo{author}{Meng, K.}, \bibinfo{author}{Bau, D.},
  \bibinfo{author}{Andonian, A.}, \bibinfo{author}{Belinkov, Y.},
  \bibinfo{year}{2022}.
\newblock \bibinfo{title}{Locating and editing factual associations in gpt},
  in: \bibinfo{editor}{Koyejo, S.}, \bibinfo{editor}{Mohamed, S.},
  \bibinfo{editor}{Agarwal, A.}, \bibinfo{editor}{Belgrave, D.},
  \bibinfo{editor}{Cho, K.}, \bibinfo{editor}{Oh, A.} (Eds.),
  \bibinfo{booktitle}{Advances in Neural Information Processing Systems},
  \bibinfo{publisher}{Curran Associates, Inc.}. pp.
  \bibinfo{pages}{17359--17372}.
\newblock \URLprefix
  \url{https://proceedings.neurips.cc/paper_files/paper/2022/file/6f1d43d5a82a37e89b0665b33bf3a182-Paper-Conference.pdf}.
\bibitem[{Mirzadeh et~al.(2021)Mirzadeh, Farajtabar, Gorur, Pascanu and
  Ghasemzadeh}]{mirzadeh2021linear}
\bibinfo{author}{Mirzadeh, S.I.}, \bibinfo{author}{Farajtabar, M.},
  \bibinfo{author}{Gorur, D.}, \bibinfo{author}{Pascanu, R.},
  \bibinfo{author}{Ghasemzadeh, H.}, \bibinfo{year}{2021}.
\newblock \bibinfo{title}{Linear mode connectivity in multitask and continual
  learning}, in: \bibinfo{booktitle}{International Conference on Learning
  Representations}.
\newblock \URLprefix \url{https://openreview.net/forum?id=Fmg_fQYUejf}.
\bibitem[{Singh and Jaggi(2020)}]{NEURIPS2020_Singh_Jaggi}
\bibinfo{author}{Singh, S.P.}, \bibinfo{author}{Jaggi, M.},
  \bibinfo{year}{2020}.
\newblock \bibinfo{title}{Model fusion via optimal transport}, in:
  \bibinfo{editor}{Larochelle, H.}, \bibinfo{editor}{Ranzato, M.},
  \bibinfo{editor}{Hadsell, R.}, \bibinfo{editor}{Balcan, M.},
  \bibinfo{editor}{Lin, H.} (Eds.), \bibinfo{booktitle}{Advances in Neural
  Information Processing Systems}, \bibinfo{publisher}{Curran Associates,
  Inc.}. pp. \bibinfo{pages}{22045--22055}.
\newblock \URLprefix
  \url{https://proceedings.neurips.cc/paper_files/paper/2020/file/fb2697869f56484404c8ceee2985b01d-Paper.pdf}.

\end{thebibliography}

\newpage
\appendix

\section{Training details}
\label{appendix:training_details}

\paragraph{Tiny-10} Tiny-10 is a simple non-residual convolutional network where convolutional, normalization, and activation layers are following each other. Table~\ref{table:tiny-10} contains the precise architecture of the variant we were using in our experiments. We used Layer Normalization \citep{ba2016layer} to follow the methodology of \cite{ainsworth2022rebasin}.

We trained our models for 300 epochs, we used SGD with momentum $0.9$. The initial learning rate was set to $0.01$, which was then divided by 10 at the third and the two thirds of the training. The batch size was $100$. We used weight decay with value $10^{-4}$. 

\begin{table}[h]
\caption{The Tiny-10 architecture we used.}
\label{table:tiny-10}
\centering
\begin{tabular}{l l}
\toprule
\multicolumn{1}{c}{Tiny-10}\\
\midrule
Layers & Name \\
\midrule
3 × 3 conv. 16-LN-ReLu & Layer 1 \\
3 × 3 conv. 16-LN-ReLu  & Layer 2 \\
3 × 3 conv. 32-LN-ReLu & Layer 3\\
3 × 3 conv. 32-LN-ReLu  & Layer 4 \\
3 × 3 conv. 32-LN-ReLu   & Layer 5 \\
3 × 3 conv. 64-LN-ReLu  & Layer 6 \\
3 × 3 conv. 64-LN-ReLu  & Layer 7 \\
1 × 1 conv. 64-LN-ReLu  & Layer 8 \\
Global average pooling  &  \\
Dense & \\
Logits & \\
\bottomrule
\end{tabular}
\end{table}

\paragraph{ResNet-20} We used the ResNet-20 architecture of \citep{he2016deep}, but with LayerNorms instead of BatchNorms. 

We trained our models for 250 epochs, we used SGD with momentum $0.9$. We applied a weight decay regularization term of $10^{-4}$, and used cosine decay schedule with linear warmup: the learning rate was initialized at $10^{-6}$ and linearly increased to $10^{-1}$ over an epoch and after a cosine decay schedule was used for the rest of the training. The batch size was $100$. 

\paragraph{Data augmentation}

The following data augmentation was performed during training for both models: 
\begin{itemize}[noitemsep]
    \item Random resizes of the image with scale factors between $0.8$ and $1.2$.
    \item Random $32\times 32$ pixel crops.
    \item Random horizontal flips.
    \item Random rotations between $\pm 30^\circ$.
\end{itemize}

\newpage
\section{Additional figures}
\label{appendix:additional_figures}

All our experiments were conducted on both the ResNet-20 and Tiny-10 models. In this section, we present all model combination plots for Tiny-10. 

\subsection{Sampling from the unit hypercube}

\begin{figure}[H]
\centering
    \subfloat[Loss]{%
        \includegraphics[trim=0cm 0cm 0cm 0cm, clip=true, width=0.5\textwidth]{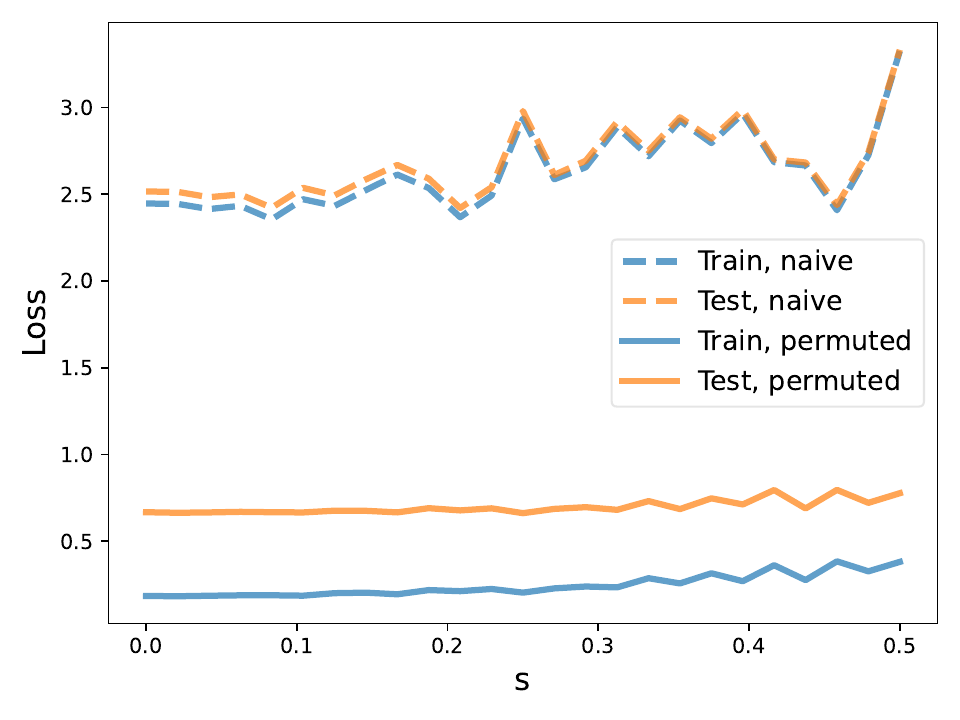}%
        \label{fig:uniform_tiny_loss}%
        }%
    \hfill%
    \subfloat[Accuracy]{%
        \includegraphics[trim=0cm 0cm 0cm 0cm, clip=true, width=0.5\textwidth]{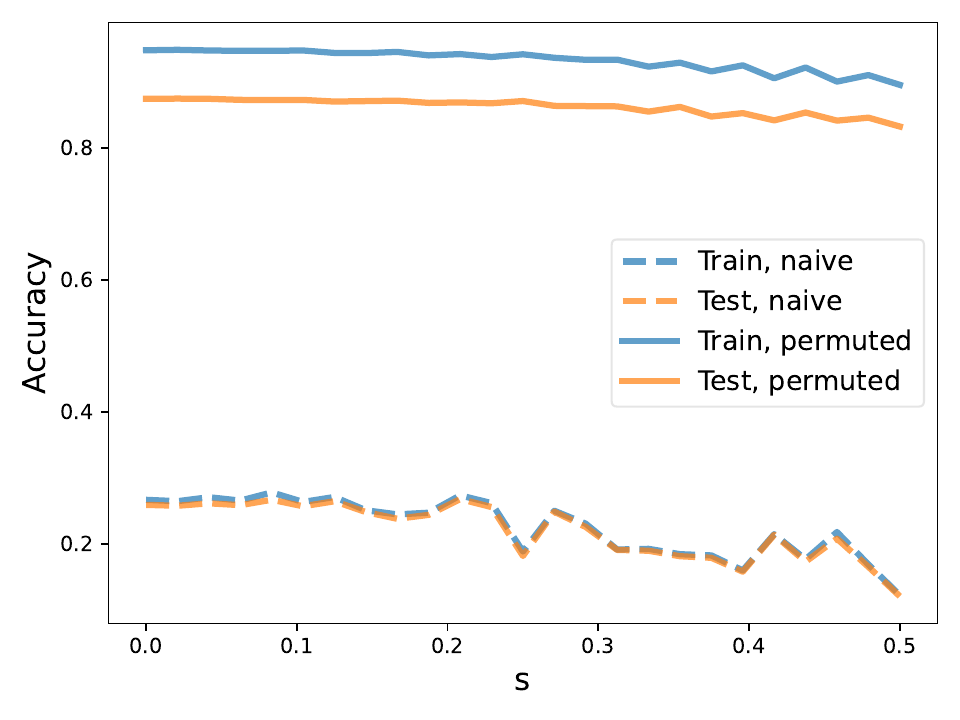}%
        \label{fig:uniform_tiny_acc}%
        }%
    \caption{Performance of Tiny-10 model combinations corresponding to uniform sampling of element-wise coefficients between $[0.5-s, 0.5+s]$. Horizontal axes denote $s$ (with $25$ values chosen equidistantly), vertical axes denote performance in terms of (a) loss and (b) accuracy. (Empirical loss barrier: 0.332, empirical accuracy barrier: 0.07.)}%
    \label{fig:uniform_tiny}%
\end{figure}%

\begin{figure}[H]
        \centering
        \subfloat[Loss]{%
            \includegraphics[trim=0cm 0cm 0cm 0cm, clip=true, width=0.5\textwidth]{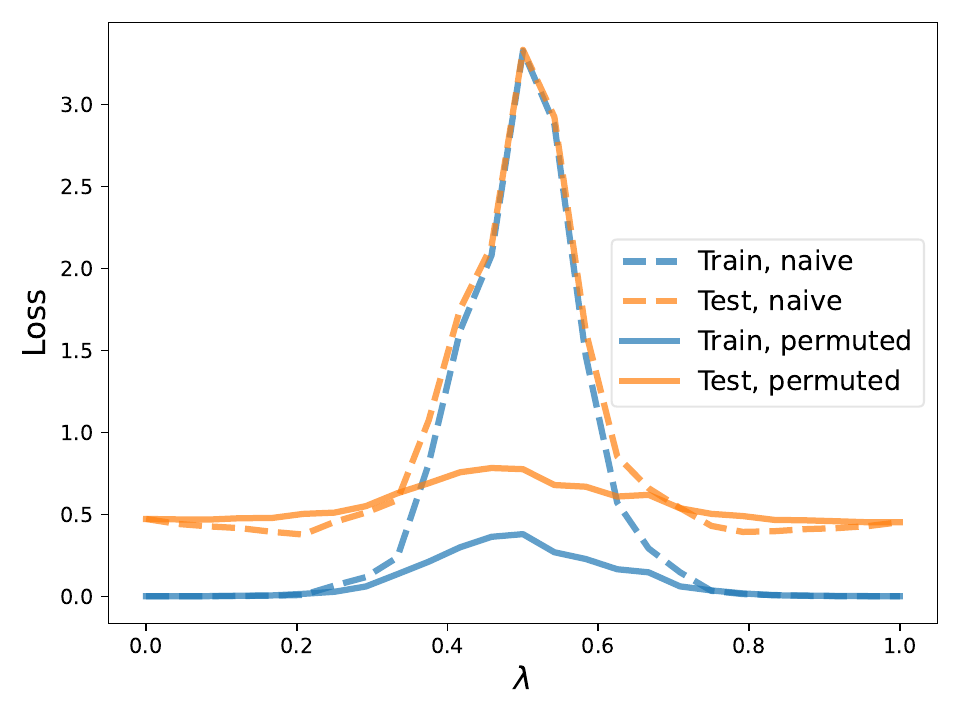}%
            \label{fig:cube_intersect_tiny_loss}%
            }%
        \hfill%
        \subfloat[Accuracy]{%
            \includegraphics[trim=0cm 0cm 0cm 0cm, clip=true, width=0.5\textwidth]{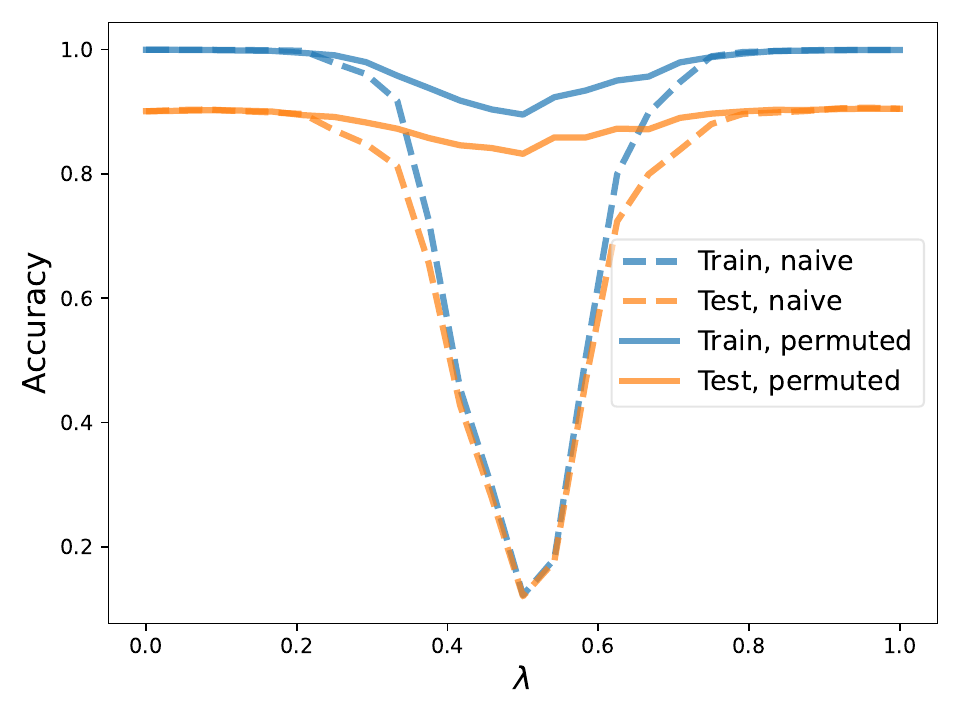}%
            \label{fig:cube_intersect_tiny_acc}%
            }%
        \caption{Performance of Tiny-10 model combinations with coefficients sampled from smaller cubes as described in Section~\ref{sec:sampling}. Horizontal axes denote the center of the small hypercube (with $25$ values chosen equidistantly), vertical axes denote performance in terms of (a) loss and (b) accuracy. (Empirical loss barrier: 0.32, empirical accuracy barrier: 0.07.)}
        \label{fig:cube_intersect_tiny}
\end{figure}

\begin{figure}[H]
    \centering
    \subfloat[Loss]{%
        \includegraphics[trim=0cm 0cm 0cm 0cm, clip=true, width=0.5\textwidth]{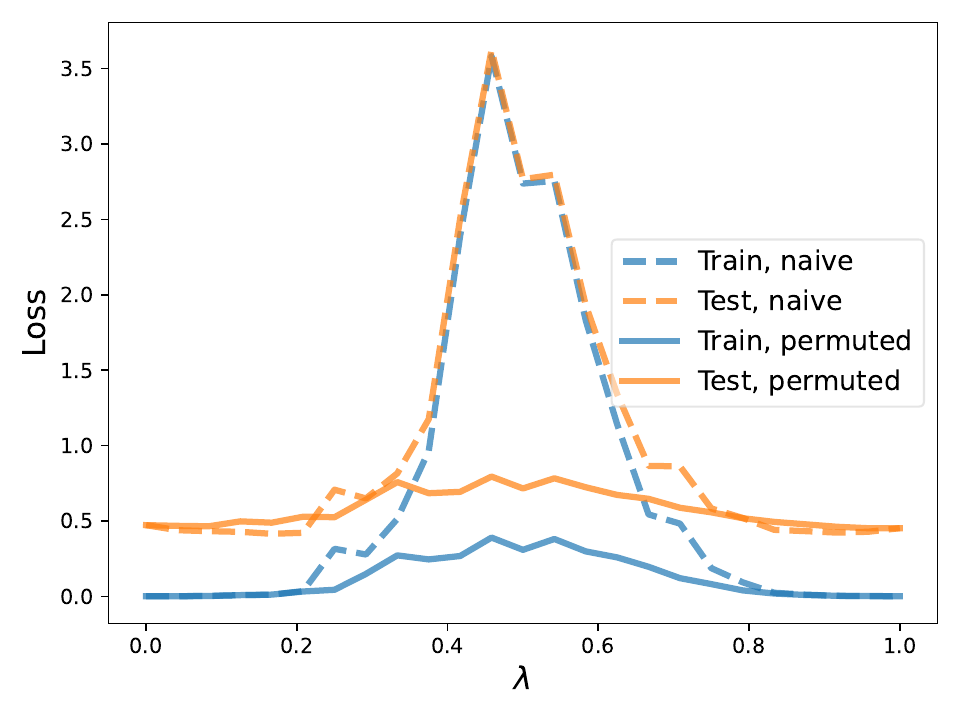}%
        \label{fig:plane_intersect_tiny_loss}%
        }%
    \hfill%
    \subfloat[Accuracy]{%
        \includegraphics[trim=0cm 0cm 0cm 0cm, clip=true, width=0.5\textwidth]{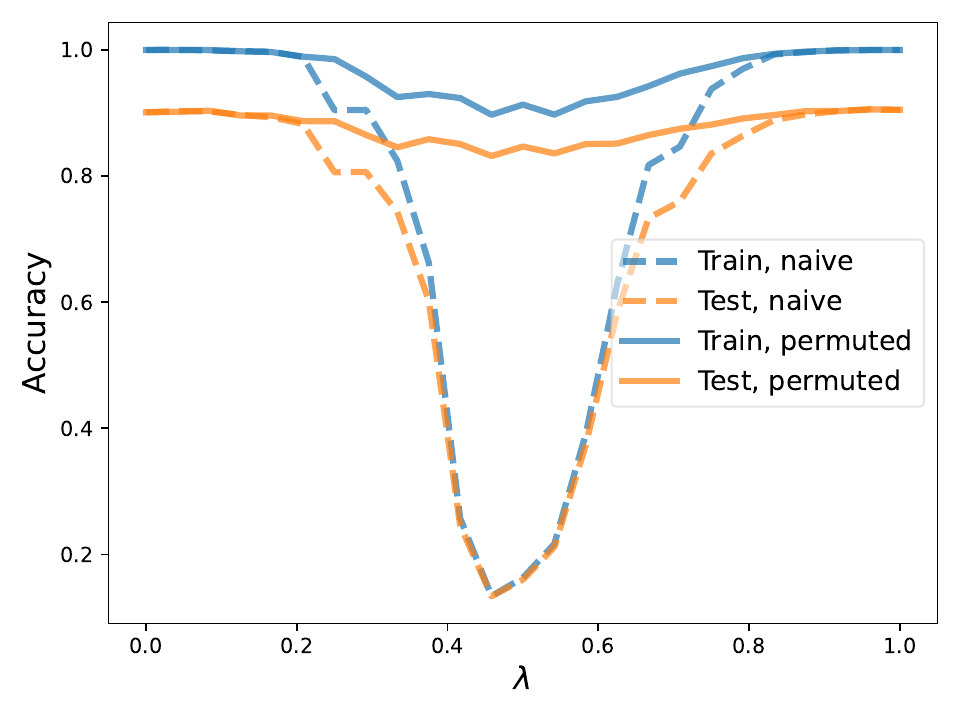}%
        \label{fig:plane_intersect_tiny_acc}%
        }%
    \caption{Performance of Tiny-10 model combinations for uniform distribution on the hyperplane described in Section~\ref{sec:sampling}. (Empirical loss barrier: 0.332, empirical accuracy barrier: 0.071.)}
    \label{fig:plane_intersect_tiny}
\end{figure}

\begin{figure}[H]
    \centering
    \subfloat[Loss]{%
        \includegraphics[trim=0cm 0cm 0cm 0cm, clip=true, width=0.5\textwidth]{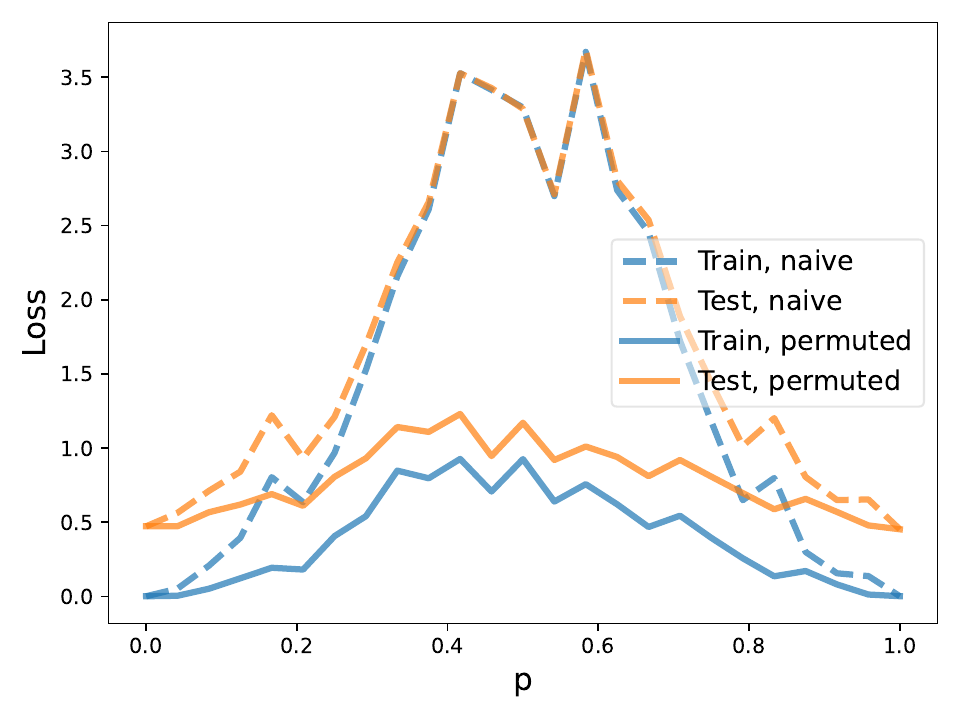}%
        \label{fig:bernoulli_tiny_loss}%
        }%
    \hfill%
    \subfloat[Accuracy]{%
        \includegraphics[trim=0cm 0cm 0cm 0cm, clip=true, width=0.5\textwidth]{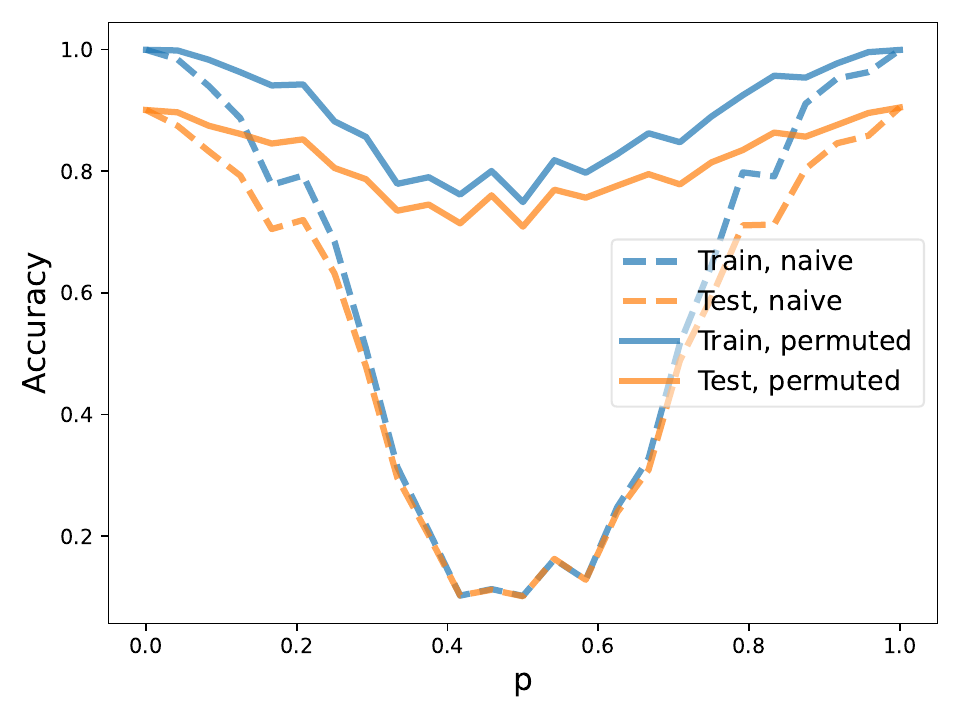}%
        \label{fig:bernoulli_tiny_acc}%
        }%
    \caption{Performance of Tiny-10 model combinations for the Bernoulli distribution on the cube. Horizontal axes denote the Bernoulli parameter $p$ (with $25$ values chosen equidistantly), vertical axes  denote performance in terms of (a) loss and (b) accuracy. (Empirical loss barrier: 0.767, empirical accuracy barrier: 0.194.)}
    \label{fig:bernoulli_tiny}
\end{figure}

\subsection{Identity stitching}

\begin{figure}[H]
\centering
    \subfloat[Loss]{
        \includegraphics[trim=0cm 0cm 0cm 0cm, clip=true, width=0.49\linewidth]{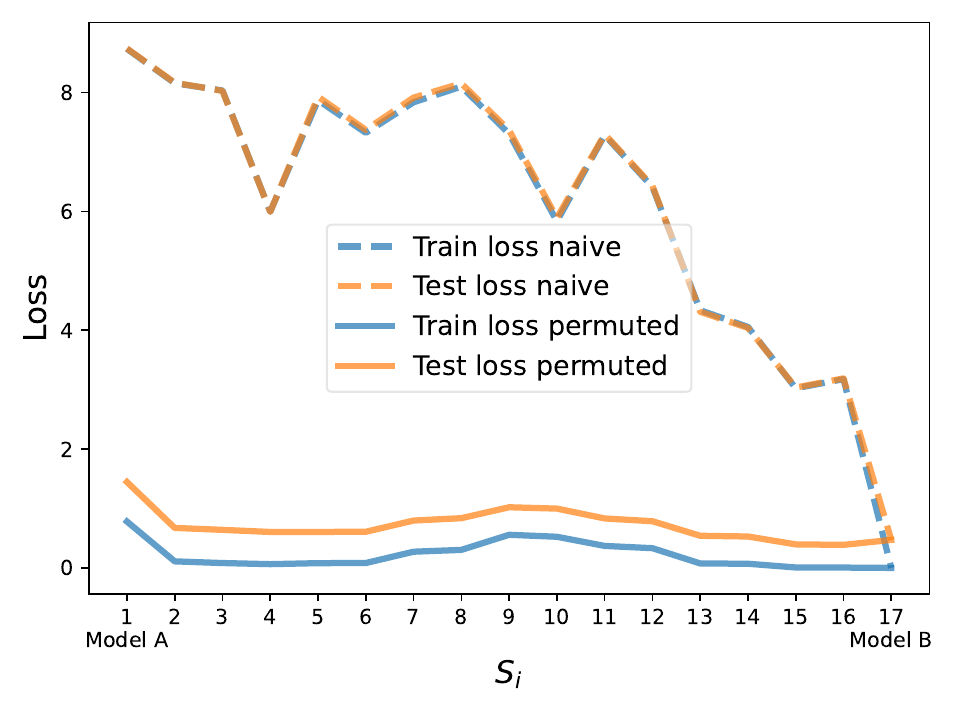}
    }%
    \subfloat[Accuracy]{
        \includegraphics[trim=0cm 0cm 0cm 0cm, clip=true, width=0.49\linewidth]{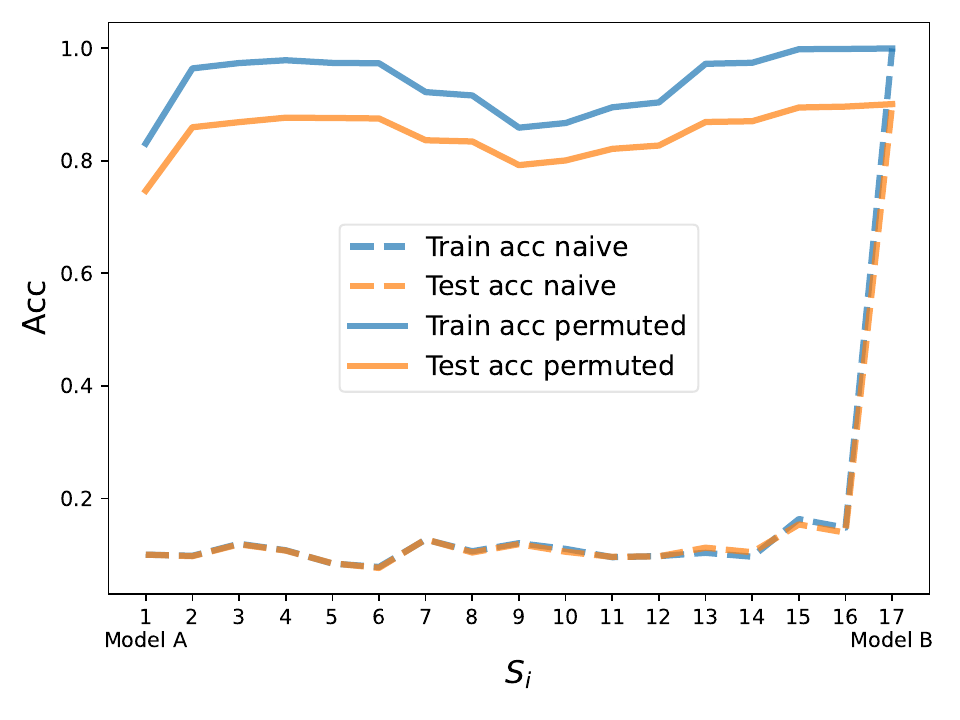}
    }
    \caption{Tiny-10 identity stitching. Plotting performance in terms of (a) loss and (b) accuracy. Horizontal axes denote the layers of the model.}
    \label{fig:stitching_tiny}
\end{figure}%

\subsection{Transitivity}

\begin{figure}[H]
    \centering
    \subfloat{%
        \centering
        \includegraphics[trim=0cm 0cm 0cm 0cm, clip=true, width=0.5\textwidth]{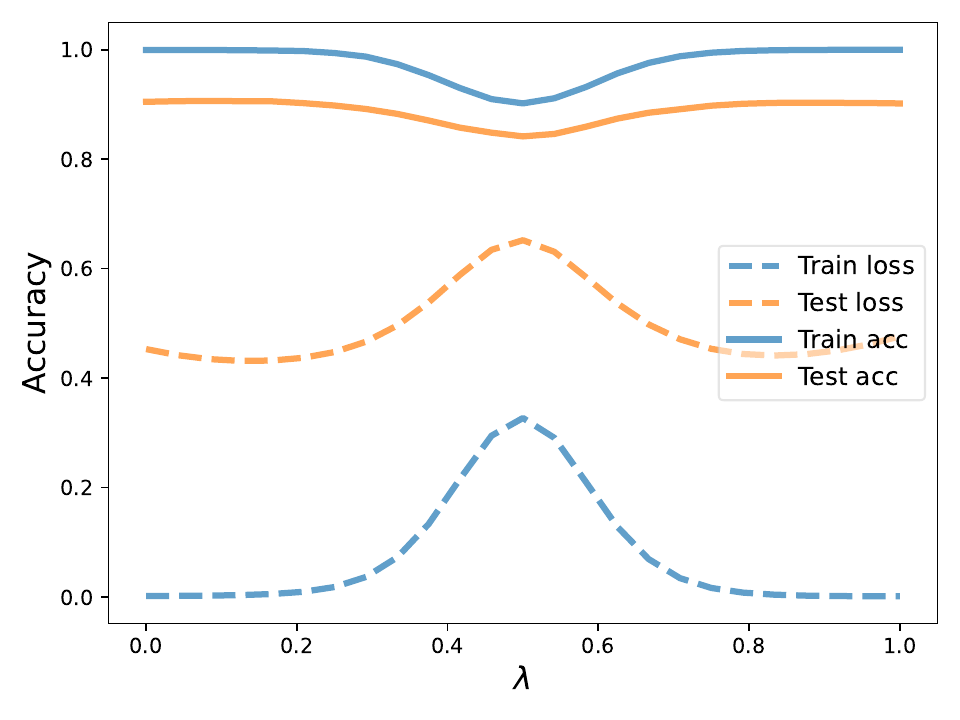}%
    }%
    \caption{Linear interpolation between two Tiny-10 models --- $\pi_B(\Theta_B)$ and model $\pi_C(\Theta_C)$ --- re-basined to a third Tiny-10 model $\Theta_A$.}
    \label{fig:tiny10_transitivity}
\end{figure}

\subsection{Width ablation}

\begin{figure}[H]
\centering
    \subfloat[Loss]{
        \includegraphics[trim=0cm 0cm 0cm 0cm, clip=true, width=0.49\linewidth]{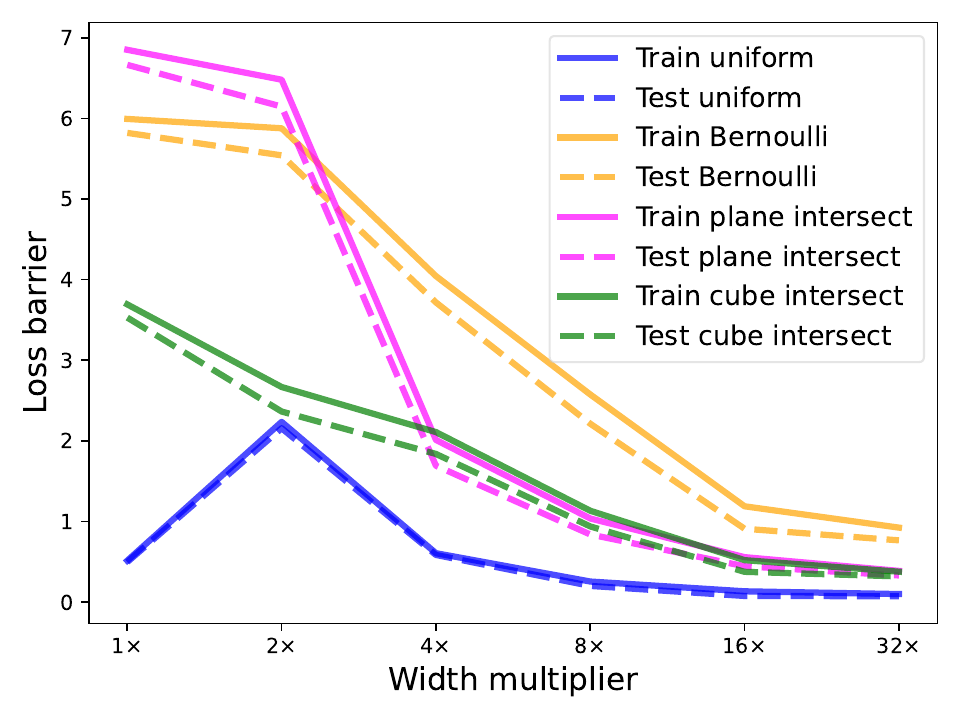}
    }%
    \subfloat[Accuracy]{
        \includegraphics[trim=0cm 0cm 0cm 0cm, clip=true, width=0.49\linewidth]{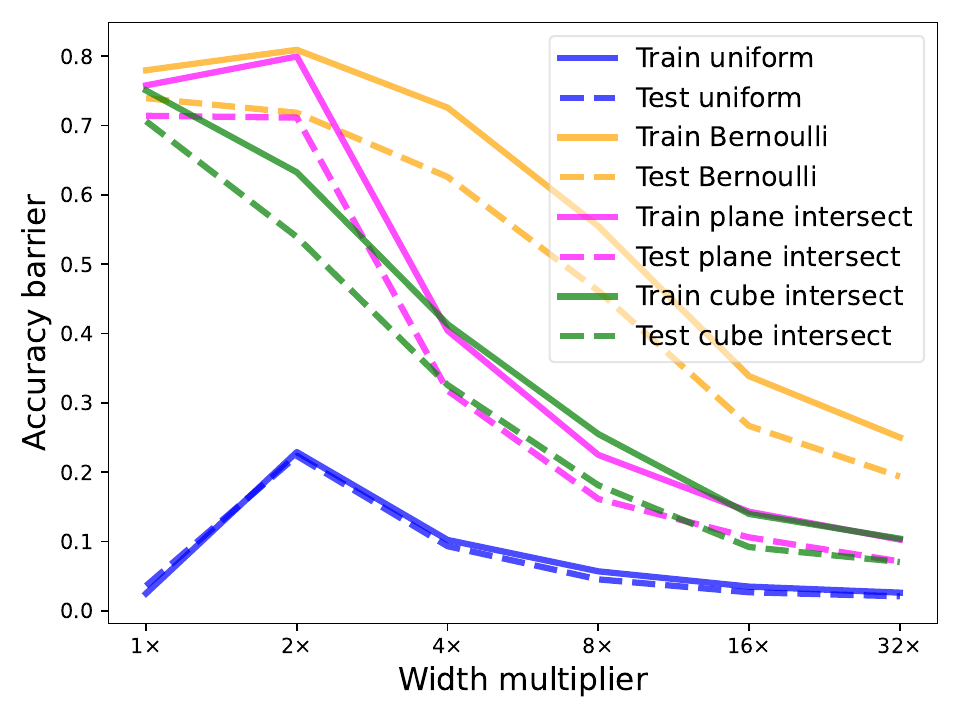}
    }
    \caption{Different model combinations and their performance with different network widths. Network is Tiny-10. Plotting performance in terms of (a) loss barrier and (b) accuracy barrier (i.e., the worst performing model is depicted for each sampling scheme and width multiplier).}
    \label{fig:tiny10_width_plots}
\end{figure}%

\subsection{Interpolation between `Min' and `Max' models}

\begin{figure}[H]
    \centering
    \subfloat[Loss]{%
        \includegraphics[trim=0cm 0cm 0cm 0cm, clip=true, width=0.5\textwidth]{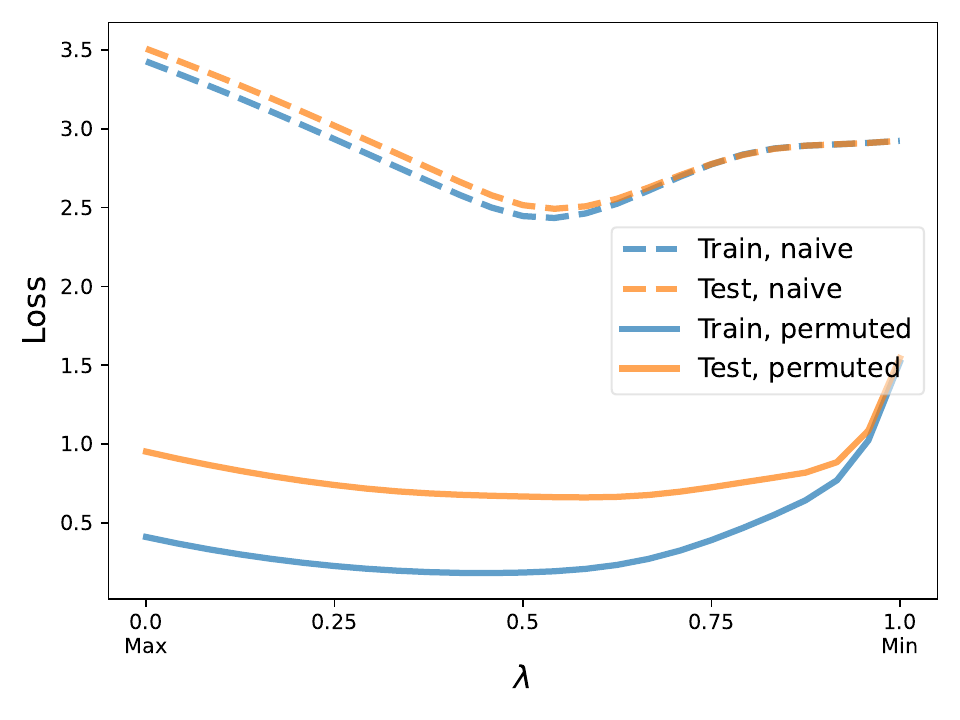}%
        \label{fig:minmax_tiny_loss}%
        }%
    \hfill%
    \subfloat[Accuracy]{%
        \includegraphics[trim=0cm 0cm 0cm 0cm, clip=true, width=0.5\textwidth]{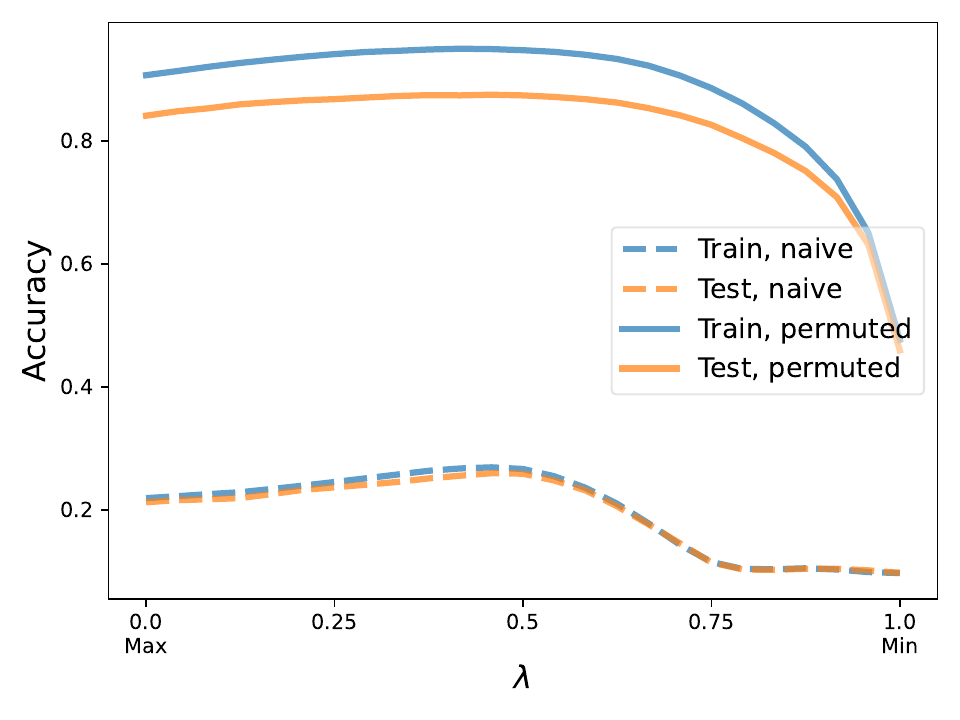}%
        \label{fig:minmax_tiny_acc}%
        }%
    \caption{Linear interpolation between the `Max' model and the `Min' model. The model was ResNet-20. Horizontal axes denote the $\lambda$ interpolation coefficient, vertical axes denote performance in terms of (a) loss and (b) accuracy.}
    \label{fig:minmax_tiny}
\end{figure}

\subsection{Extrapolation}

\begin{figure}[H]
    \centering
    \subfloat[Loss]{%
        \includegraphics[trim=0cm 0cm 0cm 0cm, clip=true, width=0.5\textwidth]{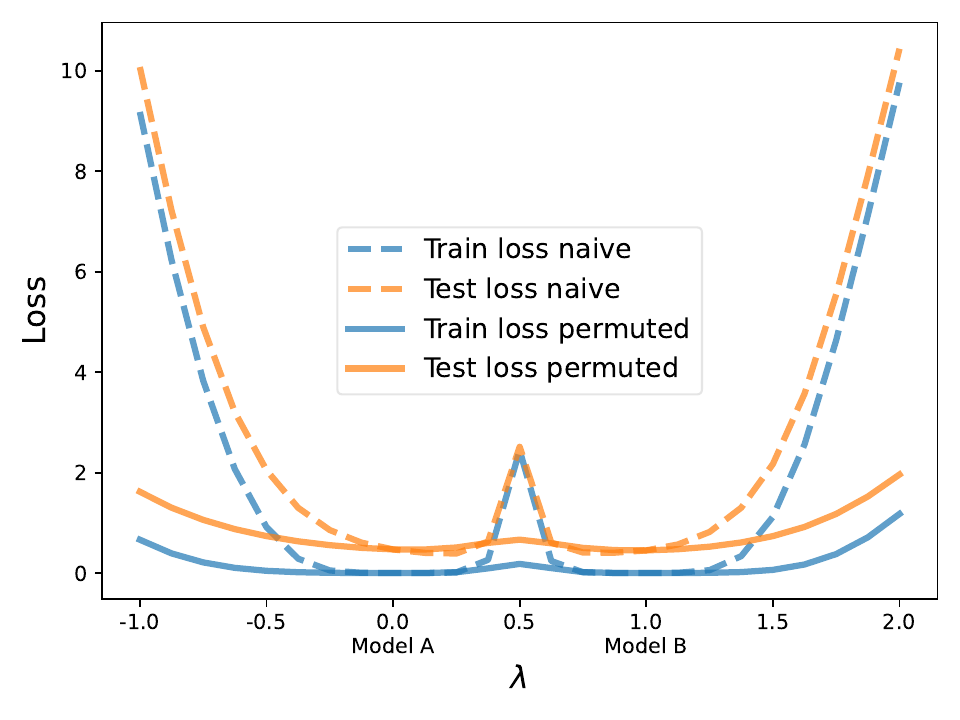}%
        \label{fig:linear_extrapolation_tiny_loss}%
        }%
    \hfill%
    \subfloat[Accuracy]{%
        \includegraphics[trim=0cm 0cm 0cm 0cm, clip=true, width=0.5\textwidth]{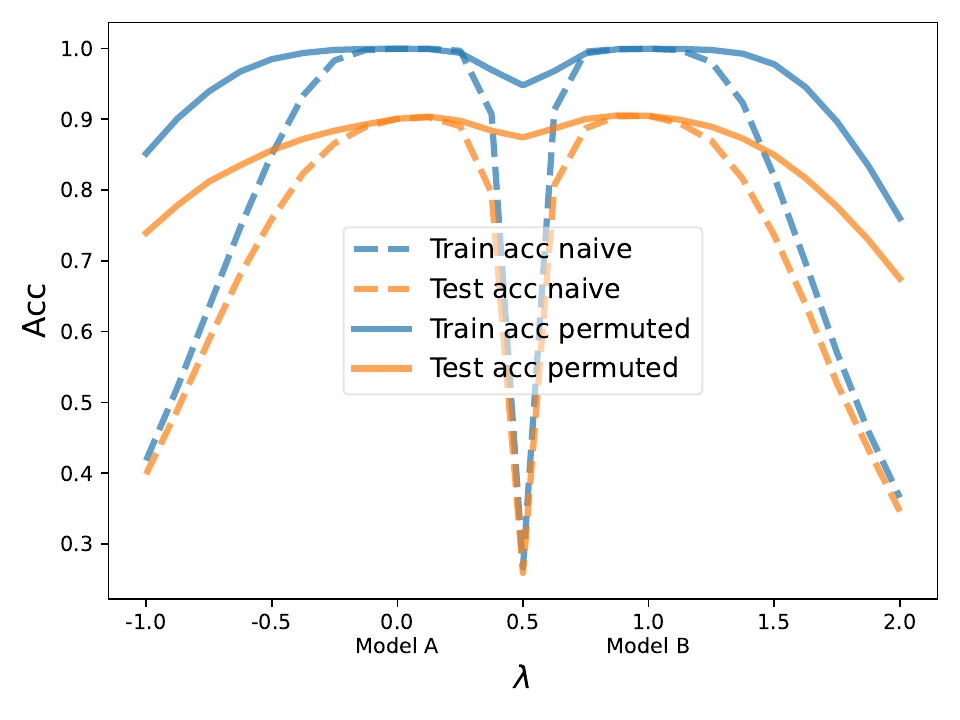}%
        \label{fig:linear_extrapolation_tiny_acc}%
        }%
    \caption{Linear extrapolation with models trained on Tiny-10. Horizontal axes denote the $\lambda$ interpolation coefficient, vertical axes denote performance in terms of (a) loss and (b) accuracy.}
    \label{fig:linear_extrapolation_tiny}
\end{figure}

\begin{figure}[H]
    \centering
    \subfloat[Train loss]{%
        \includegraphics[trim=0cm 0cm 0cm 0cm, clip=true, width=0.5\textwidth]{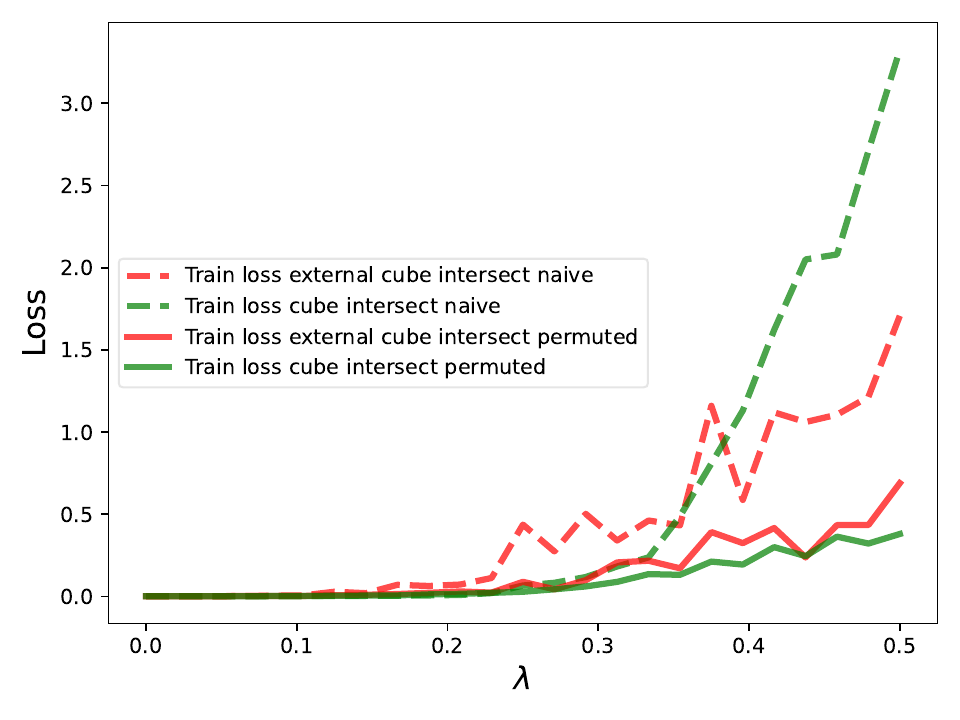}%
        \label{fig:multi_distr_extrapolation_tiny_loss}%
        }%
    \hfill%
    \subfloat[Test loss]{%
        \includegraphics[trim=0cm 0cm 0cm 0cm, clip=true, width=0.5\textwidth]{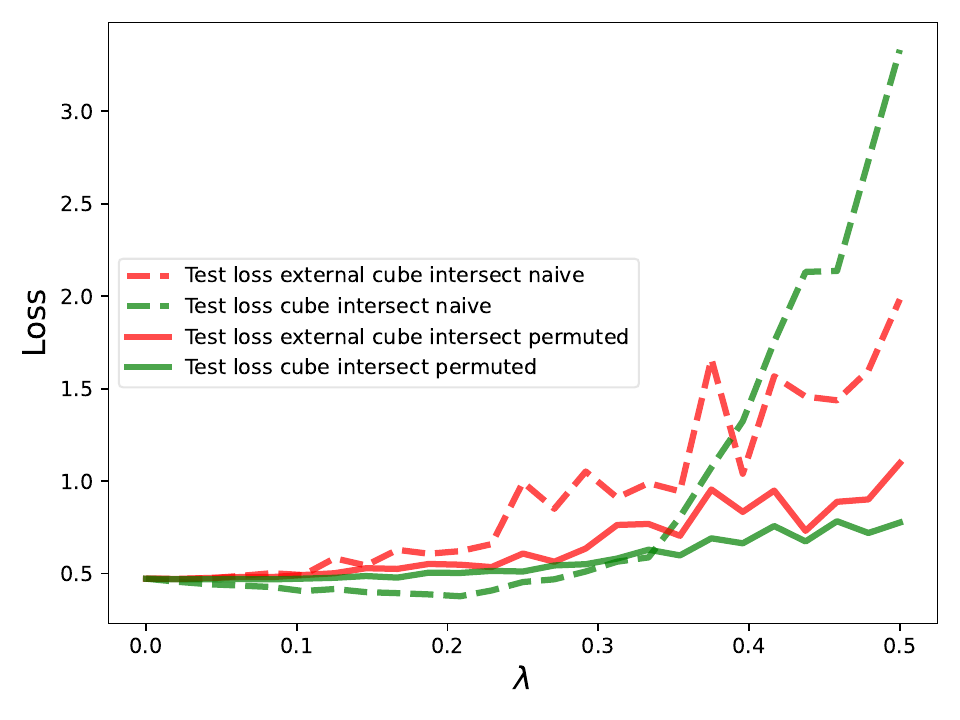}%
        \label{fig:multi_distr_extrapolation_tiny_acc}%
        }%
    \caption{Performance of Tiny-10 model combinations sampled uniformly from $\Theta_A$ centered hyperrectangles. Horizontal axis is the scaling parameter of the box, vertical axes represent performance in terms of loss.}
    \label{fig:multi_distr_extrapolation_tiny}
\end{figure}

\subsection{Stack plots}

\begin{figure}[H]
\centering
    \subfloat[Bernoulli]{%
        \includegraphics[trim=0cm 0cm 0cm 0cm, clip=true, width=0.49\textwidth]{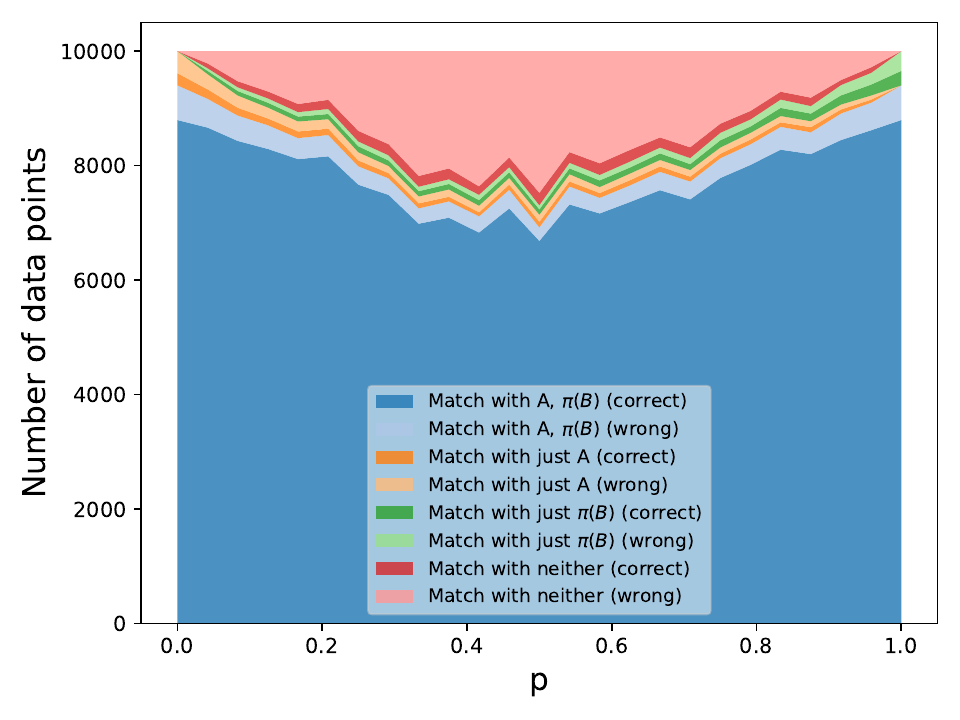}%
        \label{fig:stack_plot_bernoulli}%
        }%
    \subfloat[Cube intersect]{%
        \includegraphics[trim=0cm 0cm 0cm 0cm, clip=true, width=0.49\textwidth]{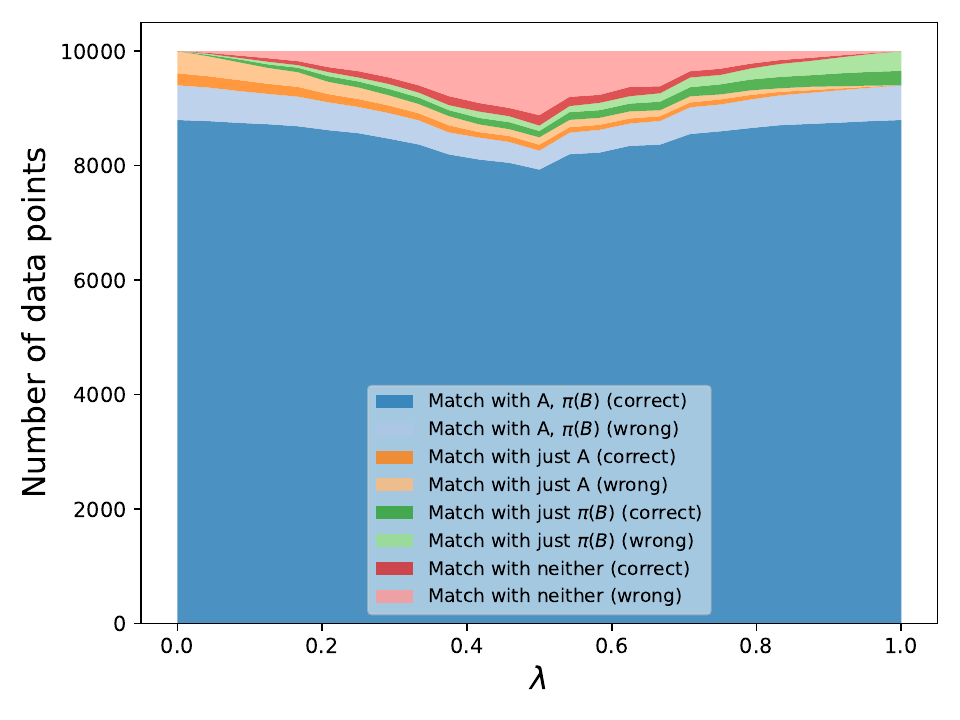}%
        \label{fig:stack_plot_cube_intersect}%
        }%
    
    \subfloat[Plane intersect]{%
        \includegraphics[trim=0cm 0cm 0cm 0cm, clip=true, width=0.49\textwidth]{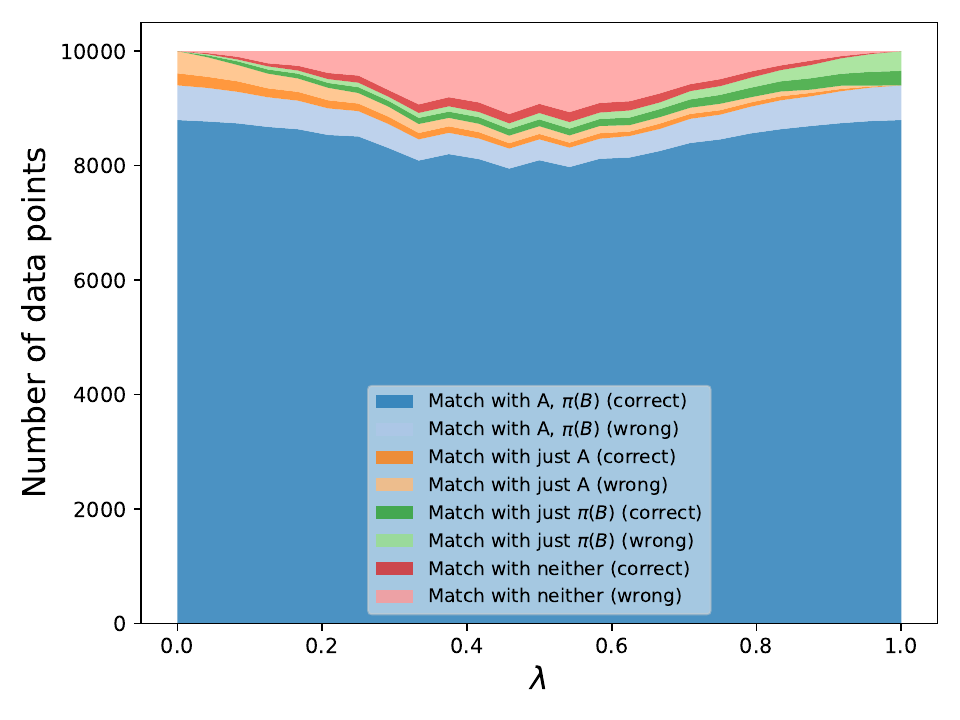}%
        \label{fig:stack_plot_plane_intersect}%
        }%
    \subfloat[Uniform]{%
        \includegraphics[trim=0cm 0cm 0cm 0cm, clip=true, width=0.49\textwidth]{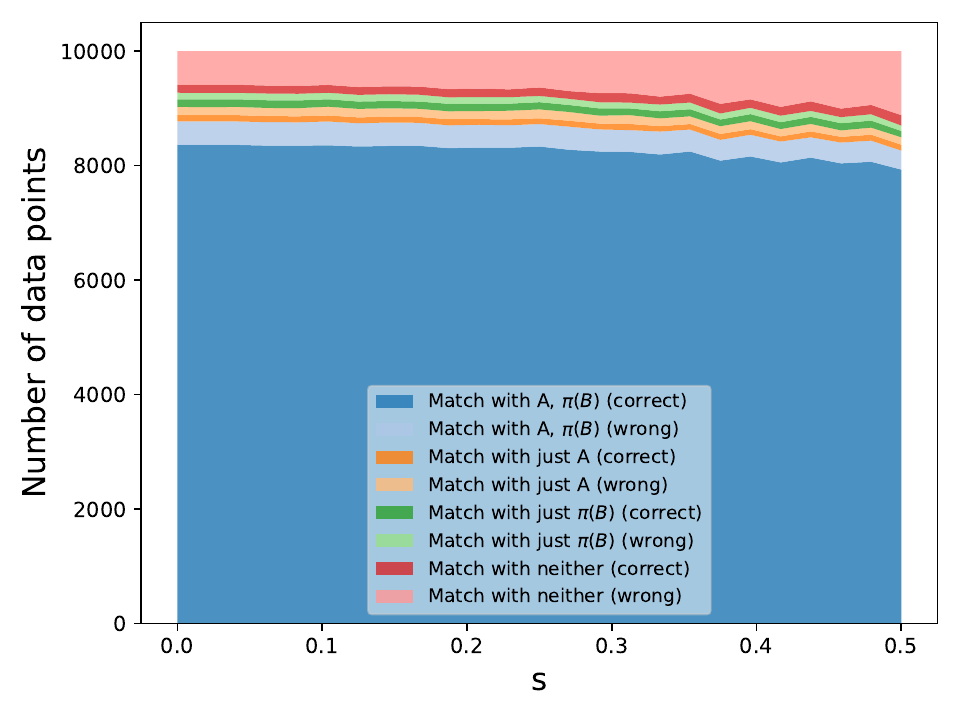}%
        \label{fig:stack_plot_uniform}%
        }%
    \caption{Network level functional comparison of Tiny-10 model combinations. The stacked plots depict how for different model combinations the 10000 datapoints of the CIFAR-10 test set are distributed among the following four categories: the predicted labels match with Model A only, Model B only, neither, or both.}
    \label{fig:stack_plots_tiny}
\end{figure}

\begin{figure}[H]
\centering
    \subfloat[ResNet-20]{%
        \includegraphics[trim=0cm 0cm 0cm 0cm, clip=true, width=0.49\textwidth]{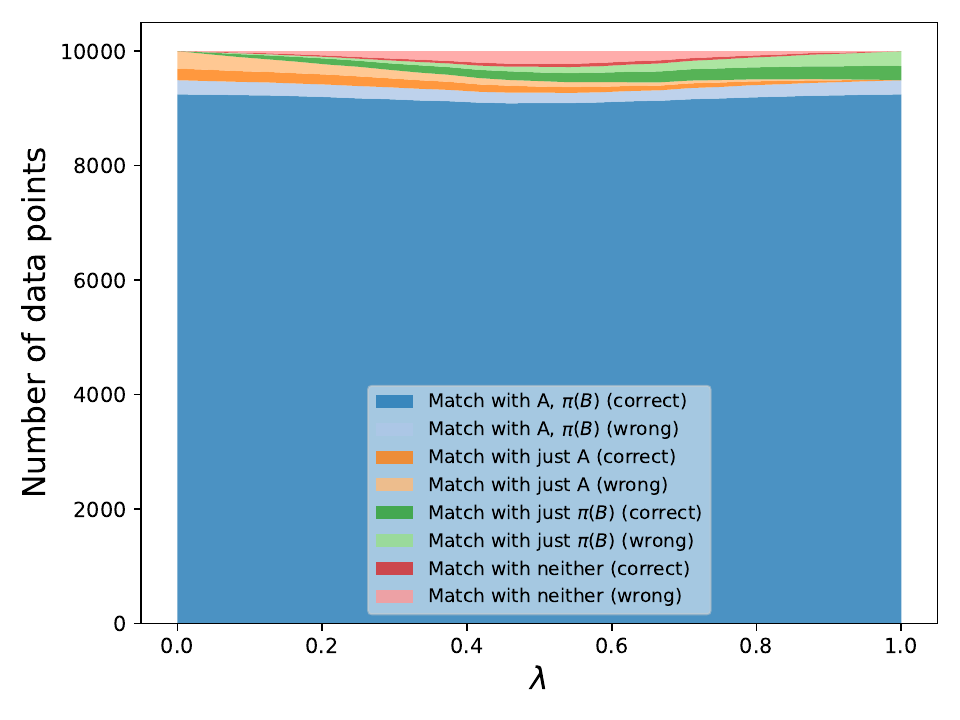}%
        \label{fig:stack_plot_lin_resnet}%
        }%
    \subfloat[Tiny-10]{%
        \includegraphics[trim=0cm 0cm 0cm 0cm, clip=true, width=0.49\textwidth]{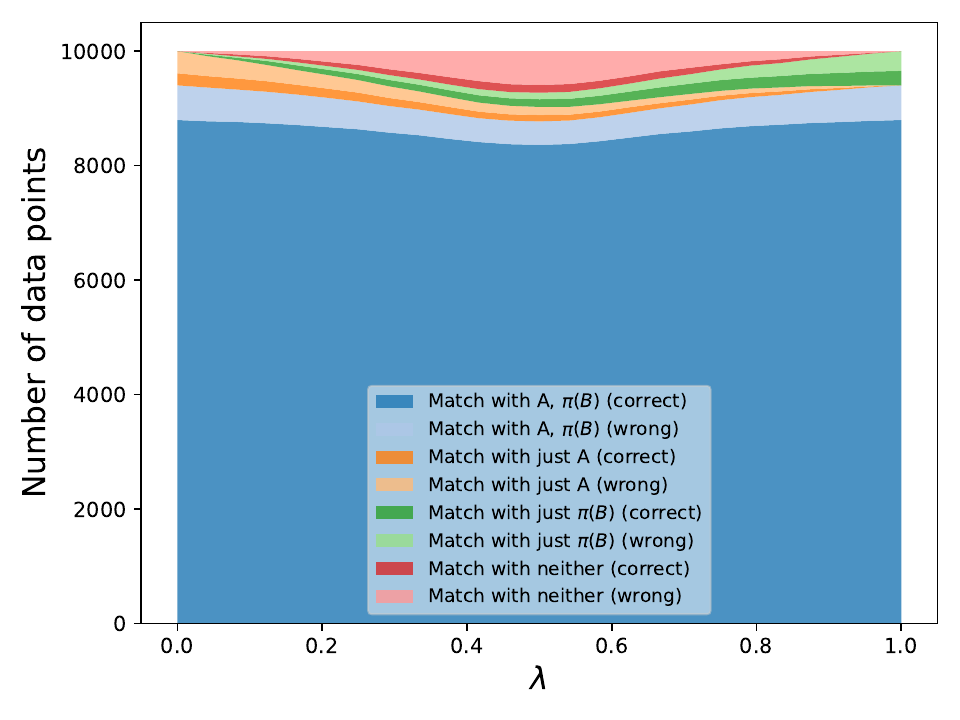}%
        \label{fig:stack_plot_lin_tiny}%
        }%
    \caption{Network level functional comparison of ResNet-20 and Tiny-10 linear interpolation. The stacked plots depict how for different model combinations the 10000 datapoints of the CIFAR-10 test set are distributed among the following four categories: the predicted labels match with Model A only, Model B only, neither, or both.}
    \label{fig:stack_plots_linear}
\end{figure}

\newpage
\section{The Re-Basin algorithm}
\label{appendix:algorithm}

For completeness, we reiterate the matching algorithm taken verbatim from \cite[Section 3]{ainsworth2022rebasin}.

\begin{algorithm}
\SetAlgoLined
\DontPrintSemicolon
\vspace{1mm}
\SetKwInOut{Input}{Given}
\Input{Model weights $\Theta_A = \left\{\mW_1^{(A)}, \dots, \mW_L^{(A)} \right\}$ and $\Theta_B = \left\{\mW_1^{(B)}, \dots, \mW_L^{(B)} \right\}$}
\vspace{1mm}
\KwResult{A permutation $\pi = \left\{ \mP_1,\dots,\mP_{L-1} \right\}$ of $\Theta_B$ such that $\vect(\Theta_A) \cdot \vect(\pi(\Theta_B))$ is approximately maximized.}
\vspace{-1mm}
\hrulefill
\vspace{1mm}

\textbf{Initialize:} $\mP_1 \gets \mI, \dots, \mP_{L-1} \gets \mI$

\Repeat{convergence}{
    \For{$\ell \in \textsc{RandomPermutation}(1,\dots,L - 1)$}{
        $\mP_\ell \gets \textsc{SolveLAP}\left(\mW_\ell^{(A)} \mP_{\ell-1} (\mW_\ell^{(B)})^\top + (\mW_{\ell+1}^{(A)})^\top \mP_{\ell+1} \mW_{\ell+1}^{(B)}\right)$
    }
}

\caption{\textsc{PermutationCoordinateDescent}}
\label{alg:permutation_coordinate_descent}
\end{algorithm}

In Algorithm~\ref{alg:permutation_coordinate_descent}, $\mW_i^{(A)}$ and $\mW_i^{(B)} \; (i \in [L])$ are the model weights corresponding to the models $\Theta_A$ and $\Theta_B$, where $L \in \mathbb{N}$ denote the number of layers. $\mP_i \; (i \in [L])$ are permutation matrices sized accordingly for the specific layers. \textsc{SolveLAP} solves the corresponding linear assignment problem (LAP) \citep{bertsekas1998network}, and \textsc{RandomPermutation} just returns a random permutation with which Algorithm~\ref{alg:permutation_coordinate_descent} overall realizes a coordinate descent running \textsc{SolveLAP}s in an iterative manner until convergence.

\end{document}